\documentclass[10pt,journal,compsoc]{IEEEtran}
\ifCLASSOPTIONcompsoc
  \usepackage[nocompress]{cite}
\else
  \usepackage{cite}
\fi
\ifCLASSINFOpdf
\else
\fi

\usepackage{amsmath,amsfonts}
\usepackage{algorithmic}
\usepackage{algorithm}
\usepackage{array}
\usepackage{textcomp}
\usepackage{stfloats}
\usepackage{url}

\usepackage{verbatim}
\usepackage{graphicx}
\usepackage{cite}
\usepackage{caption}
\usepackage{multirow}
\usepackage{color}
\usepackage{subcaption}

\usepackage{float}

\usepackage{graphicx, amsmath, amssymb, multirow, caption, overpic, textpos}

\usepackage[table]{xcolor}

\newlength\savewidth

\renewcommand{\paragraph}[1]{\vspace{1.25mm}\noindent\textbf{#1}}

\newcolumntype{x}[1]{>{\centering\arraybackslash}p{#1pt}}
\newcolumntype{y}[1]{>{\raggedright\arraybackslash}p{#1pt}}
\newcolumntype{z}[1]{>{\raggedleft\arraybackslash}p{#1pt}}

\newcommand\equalcontribution{\thanks{J. Chen and W. Luo are the co-first authors. L. Ma and W. Zhang are the corresponding authors.}}

\newcommand{\eg}[0]{\textit{e.g.}}
\newcommand{\etc}[0]{\textit{etc}}
\newcommand{\ie}[0]{\textit{i.e.}}

\usepackage{soul}
\usepackage{bbding}
\usepackage[pagebackref=true,breaklinks=true,colorlinks,bookmarks=false]{hyperref}

\begin{document}
%

%
%
%
%
\title{Learning Point-Language Hierarchical Alignment for 3D Visual Grounding}

\author{Jiaming Chen$^{\dag}$, Weixin Luo$^{\dag}$\equalcontribution, Ran Song,  Xiaolin Wei, Lin Ma$^{\S}$, and Wei Zhang$^{\S}$
\IEEEcompsocitemizethanks{\IEEEcompsocthanksitem J. Chen, R. Song and W. Zhang are with the School of Control Science and
Engineering, Shandong University, China. E-mail: ppjmchen@gmail.com, davidzhang@sdu.edu.cn.
\IEEEcompsocthanksitem W. Luo, X. Wei, and L. Ma are with Meituan, China. E-mail: forest.linma@gmail.com.
}
}

\IEEEtitleabstractindextext{
\begin{center}\setcounter{figure}{0}
    \includegraphics[width=14.2cm]{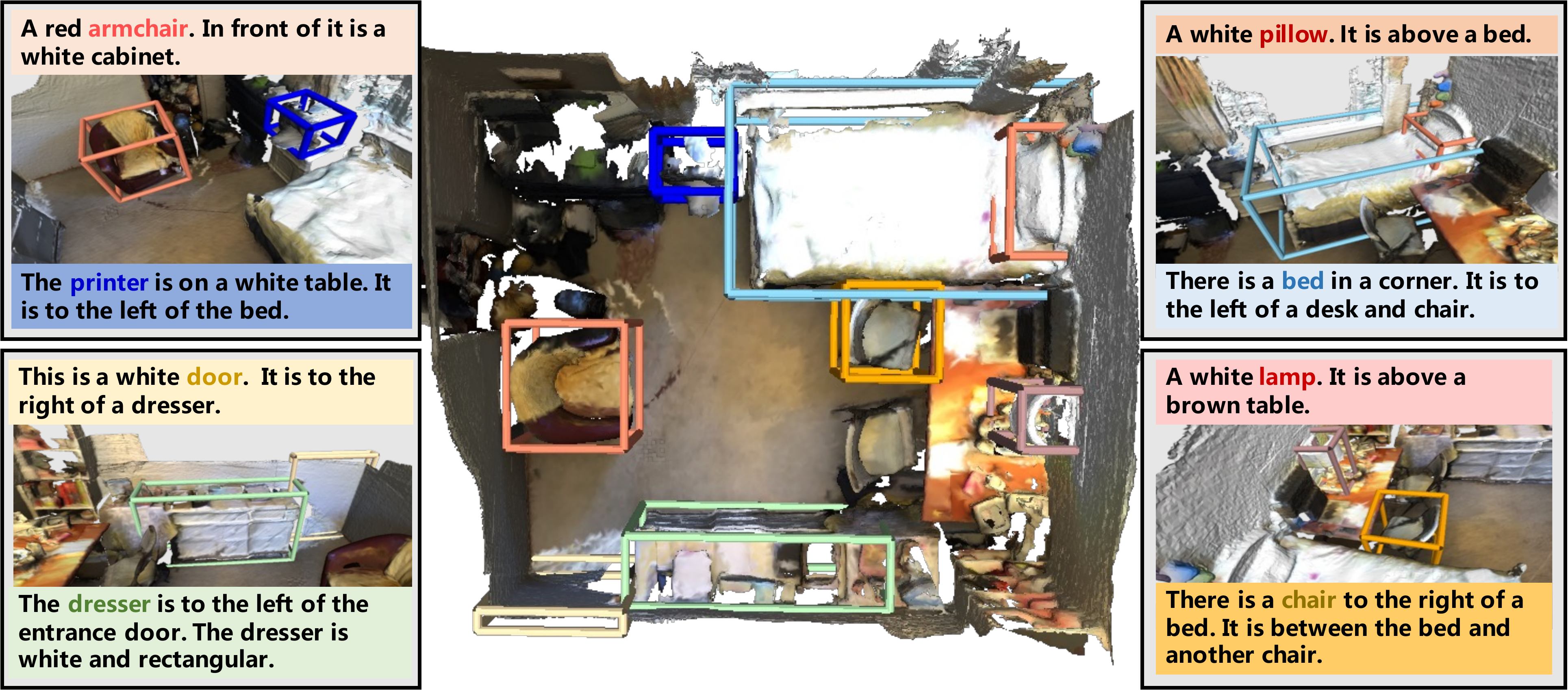}
    \captionof{figure}{\label{fig:teaser}
    Demonstration of the proposed HAM framework on the ScanRefer benchmark. This example demonstrates HAM's ability to comprehend spatial relationships through free-form language and accurately localize targets in irregular point clouds.}
\end{center}

\begin{abstract}
3D visual grounding localizes target objects on point clouds using natural language. While recent studies have made substantial progress by using Transformers to align visual and linguistic information, most of them tend to use coarse-grained attention mechanisms, which may limit their ability to comprehend lengthy and intricate language, as well as complex spatial relationships. 
This paper presents a novel hierarchical alignment model (HAM) that learns multi-granularity visual and linguistic representations in an end-to-end manner. We extract key points and proposal points to model 3D contexts and instances, and propose point-language alignment with context modulation (PLACM) mechanism, which learns to gradually align word-level and sentence-level linguistic embeddings with visual representations, while the modulation with the visual
context captures latent informative relationships. To further capture both global and local relationships, we propose a spatially multi-granular modeling scheme that applies PLACM to both global and local fields. Experimental results demonstrate the superiority of HAM, with visualized results showing that it can dynamically model fine-grained visual and linguistic representations. HAM outperforms existing methods by a significant margin and achieves state-of-the-art performance on two publicly available datasets, and won the championship in ECCV 2022 ScanRefer challenge. Code is available at~\url{https://github.com/PPjmchen/HAM}.
\end{abstract}
}
\maketitle
\IEEEdisplaynontitleabstractindextext

%
\IEEEpeerreviewmaketitle

\IEEEraisesectionheading{
\section{Introduction}\label{sec:introduction}}

\begin{figure}[t]
\centering
\includegraphics[height=3.6cm]{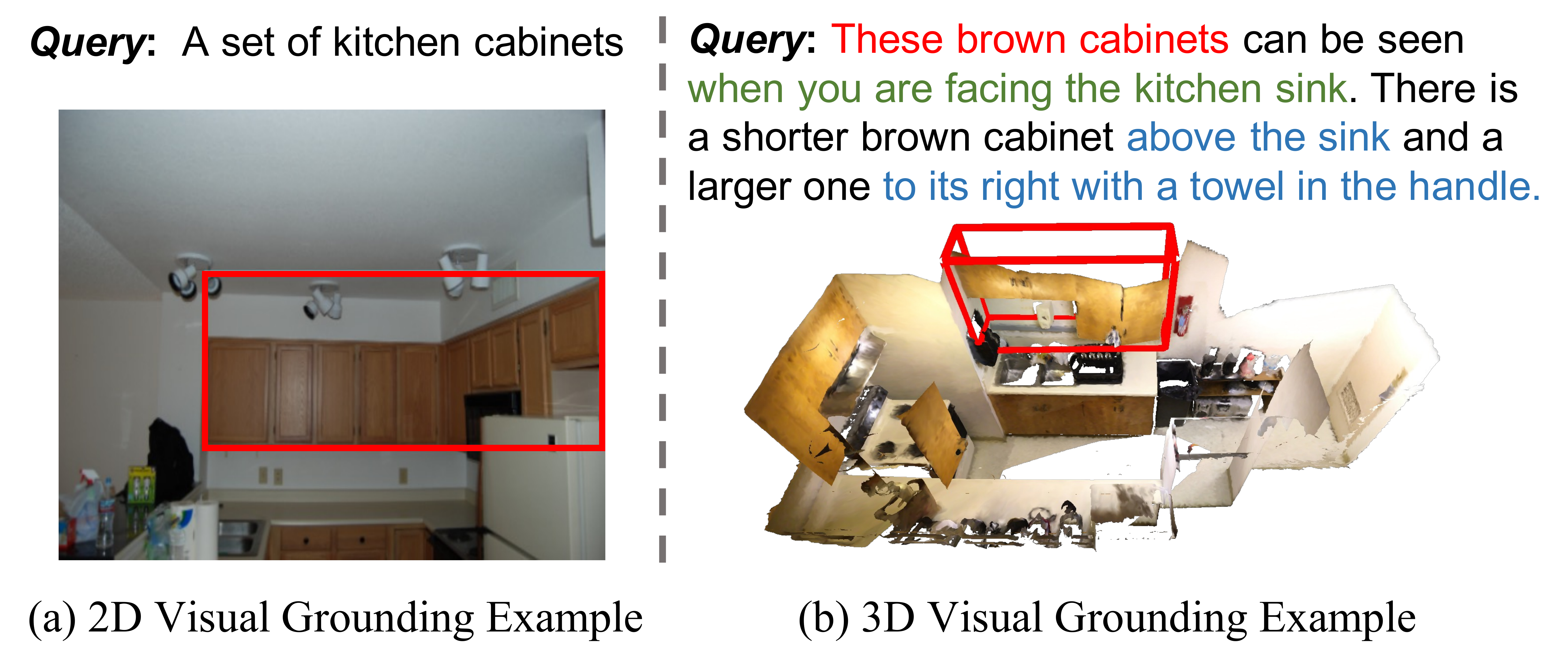}

\caption{Visualization of 2D and 3D visual grounding examples. While 2D visual grounding is typically performed on images, 3D visual grounding is more challenging, requiring a deeper understanding of the intricate spatial relationships, as well as the accompanying lengthy and complex language.}
\label{fig:compare}
\end{figure}

3D object localization on point clouds using natural language, namely 3D visual grounding, has been a prevailing research topic in multimodal 3D understanding ~\cite{scanrefer,rel3d,referit3d,refer-it-in-rgbd} since the pioneers ScanRefer~\cite{scanrefer} and ReferIt3D~\cite{referit3d} built on subsets of ScanNet~\cite{scannet}. 
Compared to the well-researched field of 2D visual grounding in images~\cite{2021mdetr,transvg,pami-grounding}, 3D visual grounding poses two significant challenges. As shown in Fig.~\ref{fig:compare}, first, 3D visual grounding requires a profound understanding of the intricate spatial relationships between 3D objects. For example, the highlighted blue phrase \textit{``above the sink''} and \textit{``to its right with a towel in the handle''}. Second, 3D visual grounding requires a comprehension of complex language queries, which may include such specifics as the observer's perspective, as highlighted in green phrase \textit{``when you are facing the kitchen sink''.}  Additionally, issues such as sensor noise, occlusion, missing data of point clouds, visual ambiguity resulting from numerous similar furniture, and other factors make it more challenging to develop accurate and robust algorithms for 3D visual grounding than for its 2D counterpart.

To tackle the two challenges, several state-of-the-art approaches~\cite{TGNN,instancerefer,ffl-3dog,3dvg-transformer,3djcg,3dsps,BUTD-DETR,languagerefer,multi-view-trans,languagerefer,sat,d3net} have been proposed, leveraging 3D point clouds abstraction networks~\cite{pointnet++}, 3D object detectors~\cite{votenet,groupfree,panoptic}, and powerful transformer networks~\cite{attention-is-all-you-need} for multimodal alignment. However, these methods often suffer common issues. First, these approaches often perform multimodal learning with coarse granularity for both language and point clouds, resulting in an insufficient understanding of lengthy and intricate language. 
Second, such methods tend to overlook the spatial context, which may lead to an incomplete grasp of complex spatial relationships. This limitation can hinder the model's ability to fully comprehend the contextual information within a scene. Addressing this issue is essential for enhancing the model's performance of visual grounding in 3D environments.

This paper proposes an end-to-end hierarchical alignment model (HAM) with multi-granularity representation learning to address the aforementioned issues. We use learnable point abstraction to gradually down-sample raw point clouds into a set of key points and further select a smaller set of proposal points to model 3D contexts and instances. The core of our method is the point-language alignment with context modulation (PLACM) mechanism, which builds upon the popular query-key-value attention~\cite{attention-is-all-you-need}. This module uses proposals as queries and combines context representations with language embeddings at both word and sentence levels to conduct hierarchical, multi-granularity attention. To capture both global and local relationships, we further propose a spatially multi-granular modeling (SMGM) scheme that applies PLACM to both global and local fields. We pre-formulate the proposal points and key points via space partition and group them regionally to obtain spatially multi-granular representations.

Given the success of prompt engineering in NLP and other multimodal domains~\cite{bert,gpt3,bridge-prompt,2021mdetr}, we hypothesize that such techniques are valuable tools for 3D visual grounding models, which rely heavily on natural language to identify and locate objects in point clouds. In this work, we extensively investigate existing prompt engineering techniques, which involve various strategies for modifying and reorganizing input text, and systematically incorporate them into our HAM framework to provide a diverse training set and enhance the efficiency and effectiveness of the training.

We evaluate the proposed HAM framework on two publicly available datasets~\cite{scanrefer,referit3d}, and show that it performs excellently in target identification and localization. Furthermore, our approach won the championship in the ECCV 2022 ScanRefer Challenge~\footnote{\url{https://kaldir.vc.in.tum.de/scanrefer\_benchmark}}. We also provide extensive ablation experiments to show the effectiveness of the proposed modules and present various visualization results that illustrate the discriminative visual and linguistic representations learned by our approach.

The main contributions of our work are summarized as follows: 
\begin{itemize}
    \item We design a hierarchical alignment model (HAM) for 3D visual grounding. It contains a point-language alignment with context modulation (PLACM) mechanism, which learns hierarchical alignment and extracts informative representations on both vision and language. A spatially multi-granular modeling (SMGM) strategy is conducted to extend PLACM to multi-granular spatial fields.
    \item We systematically analyze and incorporate three cumulative prompt engineering strategies, which enhance the performance and robustness of HAM and accelerate the training process. 
    \item Comprehensive experiments on two public datasets demonstrate the improvement of the proposed HAM. Moreover, our approach won \textbf{the championship in ECCV 2022 ScanRefer Challenge.}
\end{itemize}

We organize the rest of the manuscript as below. In Section~\ref{related_work}, we list related work, including 2D/3D visual grounding, 3D object detection, \etc. Section~\ref{approach} introduces our proposed HAM and demonstrates each module in detail. In Section~\ref{experiments}, experiments on two publicly available datasets with extensive ablation studies are carried out to validate the effectiveness of HAM. We summarize this paper in Section~\ref{conclusion}.

\section{Related Work}
\label{related_work}

Computer vision and natural language processing, two crucial sub-fields of AI, have made remarkable progress recently. Linking two modalities to realize multimodal artificial intelligence is becoming a new research trend and a broad range of related tasks has been extended. Most of the vision-language tasks can be divided into several main research fields, \eg vision-language understanding~\cite{vqa,vqav2,visual-dialog,wstg_compose,video-cap}, generation~\cite{flickr,video-cap,scan2cap,dalle,diffusion}, vision-language and robotics~\cite{vl-navigation,embodied-qa}. This work focuses on the emerging and challenging 3D visual grounding task, and we further introduce the related work in detail.

\subsection{Visual Grounding on 2D Images}

\textcolor{black}{Visual grounding was first proposed on 2D images, which }aims at localizing the target object in an image based on a given sentence query \cite{referitgame,plummer2015flickr30k,mao2016generation,yu2016modeling}. 
Most existing approaches adopt two-stage \cite{yu2018mattnet,pami-grounding,wang2019neighbourhood} and one-stage frameworks \cite{yang-osvg1,yang-osvg2,huang2021look}, which are widely used in the object detection.
Recently, a set of methods \cite{2021mdetr,clip} have made great progress in multimodal understanding tasks, benefiting from the powerful transformer network and large-scale pre-training. 
While it is tempting to leverage existing frameworks from 2D visual grounding, it is important to acknowledge that 3D visual grounding presents distinct challenges, such as irregularity, noise, and missing data in point clouds, as well as complex spatial relationships in 3D space. Consequently, developing tailored approaches that tackle these specific challenges is essential to make progress in 3D visual grounding.

\subsection{Visual Grounding on 3D Point Clouds}

Visual grounding on 3D point clouds \cite{scanrefer,referit3d} aims at localizing the corresponding 3D bounding box of a target object given query sentences and unorganized point clouds.
Two public benchmark datasets \emph{ScanRefer} \cite{scanrefer} and \emph{ReferIt3D} \cite{referit3d} are proposed. Both of them adopt the \emph{ScanNet} \cite{scannet}, an indoor scene 3D point clouds dataset, and augment it with language annotations. More specifically, ScanRefer follows the grounding-by-detection paradigm of 2D visual grounding that only raw 3D point clouds and the query sentence are given.
Alternatively, ReferIt3D formulates the 3D visual grounding as a fine-grained identification problem, which assumes the \textcolor{black}{object ground truth boxes} are known during both training and inference. 
Besides, \emph{Rel3D}~\cite{rel3d} is constructed, which focuses on grounding spatial relations between objects but not localization. \emph{SUNRefer}~\cite{refer-it-in-rgbd} is introduced for visual grounding on RGBD images. In this work, we work on the visual grounding on 3D point clouds only.

To tackle this task, most of the existing methods \cite{scanrefer,instancerefer,TGNN,referit3d,sat,3dvg-transformer,ffl-3dog} follow the two-stage framework. In the first stage, 3D object proposals are directly generated from the ground-truth \cite{referit3d} or extracted by a 3D object detector \cite{votenet}. In the second stage, the proposals and language features are aligned for semantic matching.  Recent work~\cite{sat,3dvg-transformer,transrefer3d,languagerefer} adopts the popular transformer framework to model the relationship between proposals and language powered by the attention mechanism \cite{attention-is-all-you-need}.
Some methods~\cite{TGNN,ffl-3dog,instancerefer} utilize the graph neural network to aggregate information. For instance, TGNN and InstanceRefer~\cite{TGNN,instancerefer} treat the proposals as nodes and use language to enhance them. FFL-3DOG~\cite{ffl-3dog} characterizes both the proposals and language in two independent graphs and fuses them in another graph. Recently, 3DJCG~\cite{3djcg} proposes a unified framework to jointly fit both the 3D captioning task and 3D grounding task, which consists of shared task-agnostic modules and task-specific modules. Similarly, D3Net~\cite{d3net} introduces a unified network that can Detect, Describe and Discriminate for both dense captioning and visual grounding in point clouds.  3D-SPS~\cite{3dsps} proposes a 3D single-stage referred point progressive selection method for progressively selecting key points with the guidance of language and directly locating the target in a single-stage framework. \cite{multi-view-trans} proposes a multi-view transformer for 3D visual grounding, which takes the additional multi-view features as inputs and designs the network for learning a more robust multimodal representation. BUTD-DETR~\cite{BUTD-DETR} proposes the Bottom Up Top Down DEtection TRansformers for visual grounding using language guidance and objectness guidance in both 2D and 3D.

Overall, we found that existing methods on 3D visual grounding usually perform visual-linguistic alignment upon a coarse granularity and overlook the spatial context, which inspired this work to design a hierarchical, coarse-to-fine, and multi-granularity point-language alignment framework.

\subsection{3D Object Detection} 
Since we work on end-to-end 3D visual grounding requiring accurate detection, we then revisit the related work on 3D object detection. 
To apply CNNs to 3D object detection, early works project the point clouds to the bird's view \cite{birdview1,birdview2,birdview3} or frontal views \cite{frontalview1,frontalview2}. Voxel-based methods \cite{voxel1,voxel2} conduct voxelization on point clouds and utilize 3D and 2D CNNs sequentially as used in 2D object detection. Recently, a bulk of point-based methods \cite{frustum-pointnets,pointrcnn,votenet,groupfree} put more effort into tackling this task directly on raw point clouds. Most of them group the points into object candidates by using box proposals \cite{frustum-pointnets,pointrcnn} or voting \cite{votenet}, and then extract object features from groups of points for the following detection. Group-Free \cite{groupfree} drops grounding used in voting~\cite{votenet} and uses K-Closest Point Sampling instead for proposal generation. At the same time, it applies a transformer to carry out the correlation between proposals and key points, which serves as the foundation of this paper.

More recently, RepSurf-U~\cite{RepSurf-U} presents the representative surfaces, a representation of point clouds to explicitly extract the local structure. It provides a lightweight plug-and-play backbone module for downstream tasks containing object detection. Group-Free~\cite{groupfree} equipped with RepSurf-U outperforms prior methods. 
FCAF3D~\cite{FCAF3D}, a recently proposed fully convolutional anchor-free framework for object detection in 3D indoor scenes. Different from voting-based and transformer-based methods, it provides an effective and scalable design to manipulate the voxel representation of point clouds with sparse convolutions. FCAF3D achieves the state-of-the-art 3D object detection results on three indoor large-scale 3D point clouds benchmark \textcolor{black}{datasets}~\cite{scannet,sun-rgbd,S3DIS}. In this paper, we adopt a point-based detection framework, but we believe that voxel-based design is also a strong baseline, which we leave to future work.

\subsection{Vision-Language Transformer}
Transformer-based networks demonstrate extraordinary capability in computer vision~\cite{Lian_2022_SSF, cao2022circular, qian2022svip} and natural language processing~\cite{attention-is-all-you-need,devlin2018bert,roberta} and further bridge the gap between them in the multimodality field~\cite{wstg_compose,clip,tcl,diffusion,dalle}, which also motivates us to utilize a transformer-based framework for our work. Some works~\cite{clip, tcl} conduct large-scale pre-training using a vision-language transformer. For example, CLIP~\cite{clip} introduces a transformer-based contrastive language-image pre-training method for efficiently learning visual concepts from natural language supervision. Then TCL~\cite{tcl} proposes triple contrastive learning for vision-language representation pre-training. Besides regular cross-modal alignment (CMA), TCL introduces an intra-modal contrastive (IMC) objective to provide complementary benefits in representation learning. All the network modules, including the vision encoder, text encoder, and fusion encoder, are transformer-based designs. Pre-training with two objectives, \ie, image-text matching and masked language modeling, TCL achieves state-of-the-art performance on various downstream vision-language tasks such as image-text retrieval and visual question answering.

\begin{figure*}[t]
\centering
\includegraphics[height=10cm]{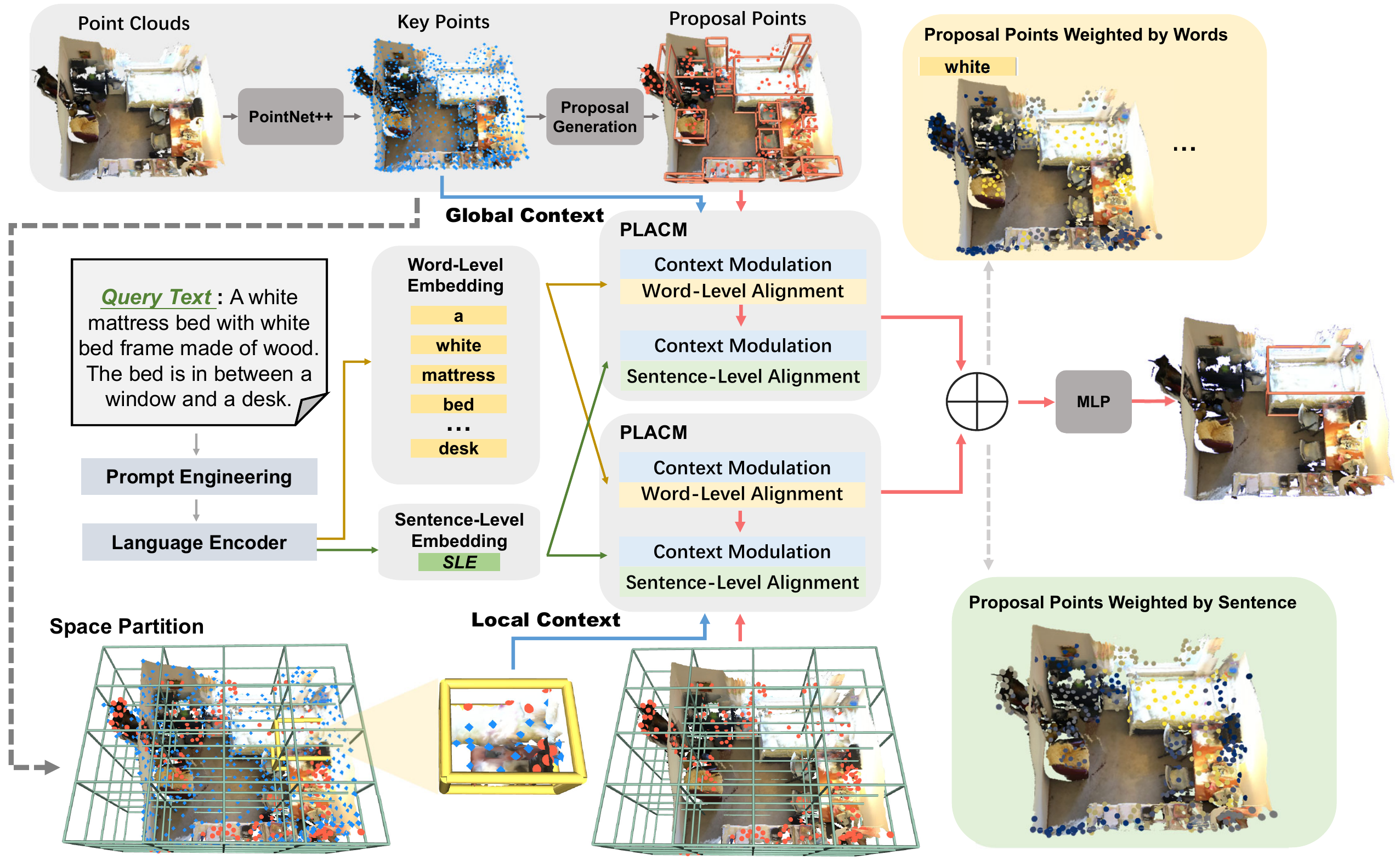}
\caption{
Workflow of the proposed hierarchical alignment model (HAM). The initial pre-processing and encoding of raw point clouds and query texts yield key points, proposal points, associated point features, proposal bounding boxes, as well as word-level and sentence-level embeddings. Two branches are developed for spatially multi-granular modeling (SMGM): one implementing point-language alignment with context modulation (PLACM) on the global field, and the other applying PLACM on local regions generated by space partitioning. The outputs of both branches are merged and passed through MLP layers to predict the final matching scores.
}
\label{fig:framework}
\end{figure*}

\section{Our Approach}
\label{approach}
Provided with a point cloud $P$ with supplementary visual features such as colors, normals, or multi-view features, and a query $Q$, the aim of 3D visual grounding is to establish a mapping $\mathcal{M}$ that links $P$ and $Q$ to the target object $o$, i.e., $\mathcal{M}(P,Q) \rightarrow o$. Essentially, this process requires a comprehensive understanding of the intricate 3D environment through the integration of visual and linguistic information.

In this paper, we design a hierarchical alignment model (HAM) to address this problem. We extract a sequence of key points and proposal points with corresponding features and proposal bounding boxes from raw point clouds. For the query texts, we perform prompt engineering to obtain diverse training samples, and a language encoder is utilized to generate both word-level and sentence-level embeddings. To achieve a comprehensive multimodal interaction, we introduce the point-language alignment with context modulation (PLACM) mechanism, which modulates proposal points using key points and promotes gradual alignment with two levels of language embeddings. We further propose the spatially multi-granular modeling (SMGM) scheme, which contains two branches to perform PLACM on both global and local fields in parallel. The outputs from both branches are merged to predict the final matching score. Grounding can then be achieved by selecting the highest matching score. We also propose an adaptation strategy for adapting the HAM to identification paradigm on two datasets~\cite{scanrefer, referit3d} with different evaluation metrics. The overall framework is illustrated in Fig.~\ref{fig:framework}.

In the following, we introduce the point cloud pre-processing and feature encoding in Section~\ref{sec:vision}. The language embedding and the prompt engineering are introduced in Section~\ref{sec:language}.  Then, the PLACM mechanism is introduced in Section~\ref{sec:placm}. In Section~\ref{sec:SMGM}, we proposed the design of SMGM scheme. The loss function and the \textcolor{black}{adaptation strategy} to the identification paradigm are introduced in Sections~\ref{sec:loss} and~\ref{sec:adapting}, respectively.

\subsection{Point Cloud Encoding}~\label{sec:vision}

\subsubsection{Key Points and Proposal Points Extraction}
Following previous methods~\cite{scanrefer,sat,3dvg-transformer,instancerefer,ffl-3dog}, we employ PointNet++~\cite{pointnet++} as the backbone network for point abstraction. Specifically, PointNet++ down-samples $L$ raw points $P\in\mathbb{R}^{L\times (3+x)}$, which includes 3D coordinates and $x$-dimensional supplementary visual features, to $N$ \textbf{key points}, denoted as $P_K=\{P_K^{(i)}\}_{i=1}^N$. Such key points sparsely represent the point cloud while capturing significant contextual information. Their corresponding features are denoted as $F_{K}=\{F_K^{(i)}\}_{i=1}^N$.

We further utilize the learnable k-closest points sampling (KPS) \cite{groupfree} on $P_K$ and $F_K$ to select a subset of $M$ \textbf{proposal points} $P_Q=\{P_Q^{(j)}\}_{j=1}^M$ and the corresponding features $F_Q=\{F_Q^{(j)}\}_{j=1}^M$, which represent the proposal objects in the point cloud. Here both $F_K$ and $F_Q$ have the same dimension $C$, \ie, $F_K\in\mathbb{R}^{N\times C}$ and $F_Q\in\mathbb{R}^{M\times C}$.
Drawing inspiration from the state-of-the-art Group-Free~\cite{groupfree} approach in 3D object detection tasks~\cite{scannet,sun-rgbd}, we employ multi-head cross-attention modules to enhance proposal points by correlating them with key points. All layers are fully supervised using the 3D object detection task on ScanNet, and we select the regressed bounding boxes at the final layer as the boxes for the proposal points. 

\subsubsection{Concentration Sampling}

Existing methods for point sampling in regular PointNet++, such as distance-FPS (D-FPS) based on 3D coordinates and feature-FPS (F-FPS), taking into account the 3D coordinate features extracted by multiple linear layers, have limitations in sampling diversity and recall of interested points. The fusion sampling (FS) employed in~\cite{3dssd}, which simply takes 50\% D-FPS samples and 50\% F-FPS samples, can lead to overlapping points and a decrease in sampling diversity.

To address these issues, we propose a simple yet effective down-sampling strategy. We name it concentration sampling (CS). To prevent overlapping points, we use a point-id queue to remove duplicates and evenly add points from D-FPS and F-FPS. Experimental results show that our proposed strategy outperforms existing down-sampling methods in terms of sampling diversity and recall of interested points, as shown in Section \ref{sec:interesting}.

\begin{figure}[t]
\centering
\includegraphics[height=5.5cm]{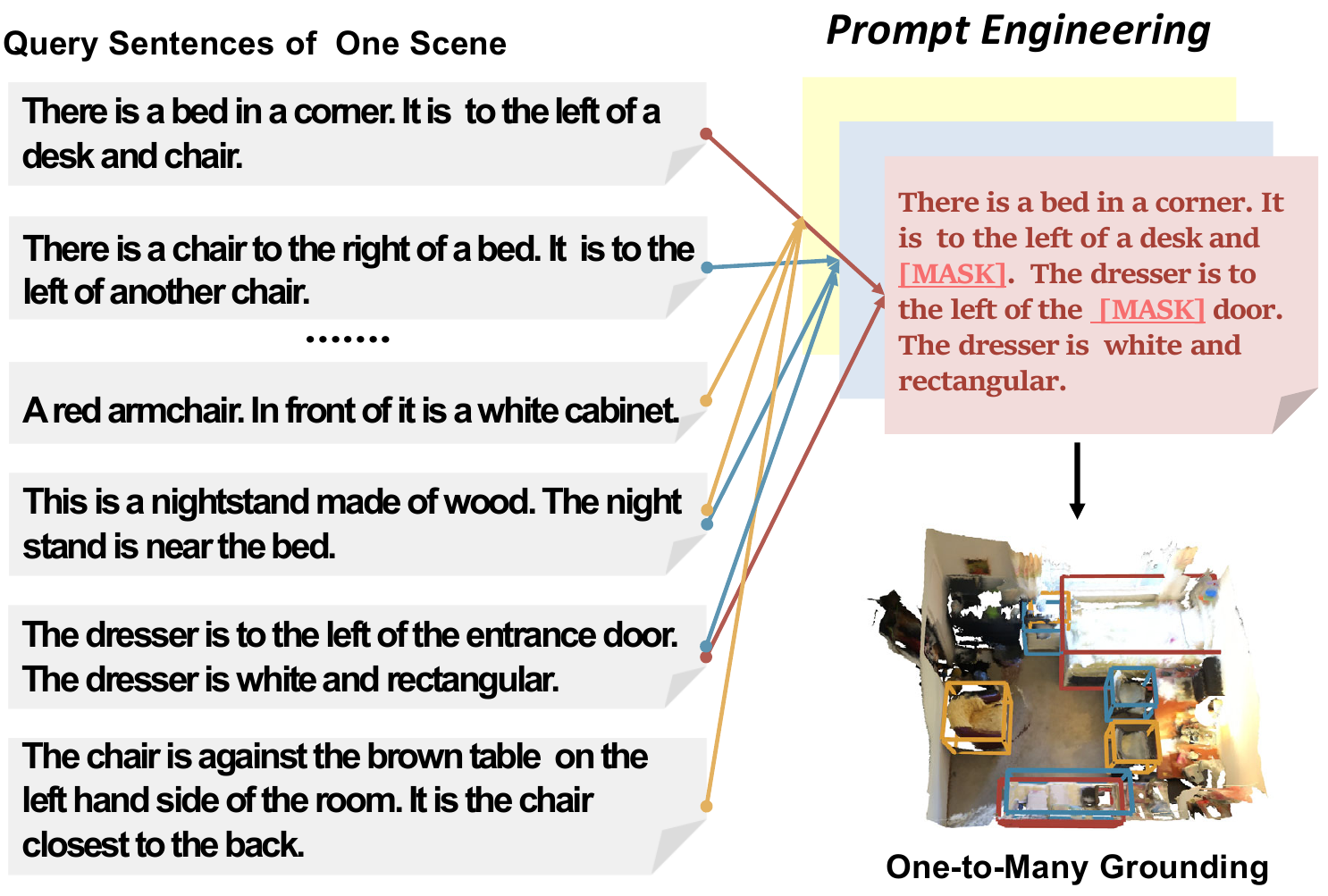}
\caption{Three cumulative prompt engineering strategies: (1) word masking, (2) intra-sentence ensemble, and (3) inter-sentence ensemble.}
\label{fig:prompt}
\end{figure}

\subsection{Language Encoding}~\label{sec:language}

\subsubsection{Multi-granularity Language Embedding}

Our hierarchical framework incorporates multi-granularity representations by embedding the natural language input query at both word and sentence levels. Such representations serve as input tokens for the PLACM module.

Specifically, to extract the initial word embeddings, we utilize a pre-trained GloVe~\cite{glove} model for a given query text. We employ a vanilla GRU~\cite{gru} to extract the \textbf{word-level embedding} $F_W\in\mathbb{R}^{T\times C}$ for each word token, where $T$ denotes the maximum length of the input text. Subsequently, the \textbf{sentence-level embedding} $F_S\in\mathbb{R}^{1\times C}$ is generated as the last-step feature of the GRU.
We opt for simple language embeddings instead of advanced text encoding techniques such as RoBERTa~\cite{roberta} in the default network setting to ensure fair comparisons and ease the end-to-end training in this work.

\subsubsection{Prompt Engineering}

To improve language modeling, we conduct a comprehensive analysis of existing prompt engineering techniques and systematically integrate them into our framework. We introduce the three cumulative strategies:

\textbf{Word Masking.} Randomly masking a portion of words from the raw text is a prevalent augmentation strategy employed in self-supervised language pre-training models, such as~\cite{bert,roberta,distilbert}, and previous multimodal transformer frameworks like ~\cite{3dvg-transformer,explore,vilt}. This approach contributes to more diverse training examples, leading to a more robust linguistic representation. In our work, we randomly mask up to $20\%$ of words in the query sentence.

\textbf{Intra-Sentence Ensemble.} To generate more complex and diverse linguistic input, we utilize an intra-sentence ensemble strategy. Inspired by MDETR~\cite{2021mdetr}, we sample multiple original queries associated with the scene and subsequently fuse them to create an ensembled sentence. This ensembled sentence refers to multiple objects within a single scene. Such an intra-sentence ensemble encourages the model to better handle complex queries and multi-object grounding tasks.

\textbf{Inter-Sentence Ensemble.} To further enhance training efficiency, recent methods~\cite{3dvg-transformer,d3net} extended the original paradigm of grounding a single sentence by grouping multiple raw sentences within a single training sample. We incorporate this strategy into the prompt engineering and name it the inter-sentence ensemble, which distinguishes it from the above intra-sentence ensemble since it results in inference with multiple individual sentences.

As illustrated in Fig.~\ref{fig:prompt}, the integration of cumulative prompt engineering strategies transforms the original sparse training sample into a one-to-many grounding manner. This results in a diverse training set, enhancing the efficiency and effectiveness of the training process, ultimately leading to more robust linguistic representations.

\begin{figure}[t]
\centering
\includegraphics[height=5.7cm]{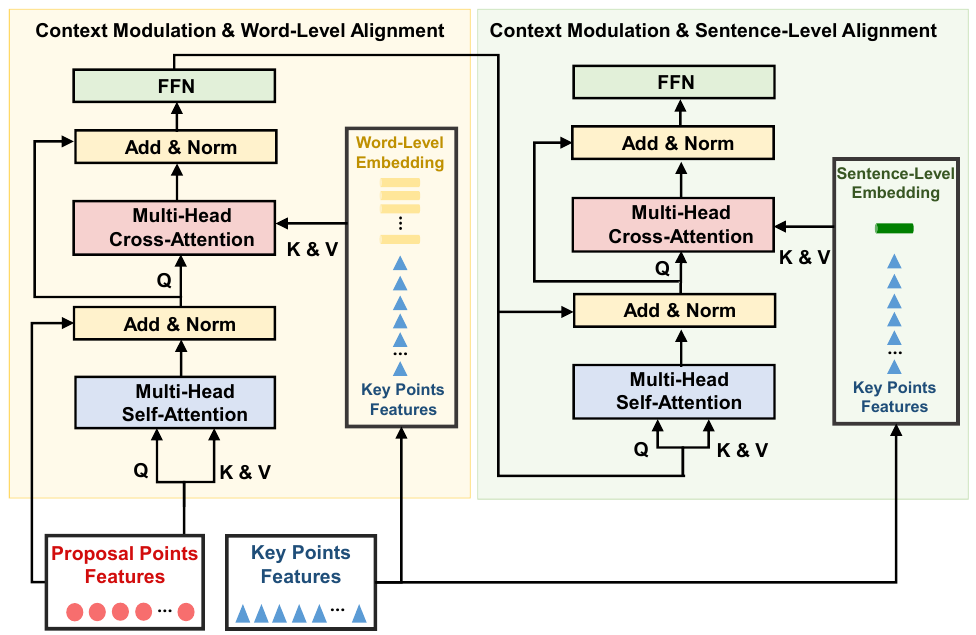}
\caption{Flowchart of the  mechanism of point-language alignment with context modulation (PLACM).}
\label{fig:placm}
\end{figure}

\subsection{PLACM: Point-Language Alignment with Context Modulation}~\label{sec:placm}

To integrate visual and linguistic features within an alignment space, we propose the mechanism of point-language alignment with context modulation (PLACM) to obtain informative language-aligned visual representations.

The PLACM is constructed with the standard multi-head self-attention (MHSA) and multi-head cross-attention (MHCA) mechanisms~\cite{attention-is-all-you-need}. Here we formally define the MHSA and the MHCA as follows:
\begin{align}
	\operatorname{MHSA}(X) &=\sigma(f_{query}(X)f_{key}(X)^{T})f_{value}(X) \\
	\operatorname{MHCA}(X, Y) &=\sigma(f_{query}(X)f_{key}(Y)^{T})f_{value}(Y)
\end{align}
where $\sigma$ is the softmax function. $f_{query}$, $f_{key}$, and $f_{value}$ represent the mapping functions of the query, key, and value, respectively.

Assuming that we have obtained a collection of proposal points and key points, along with their respective features $F_q$ and key points features $F_k$, as well as the word-level embedding $F_W$ and sentence-level embedding $F_S$ from the query text, the goal of the PLACM mechanism is to fuse and align such multimodal features effectively. This leads to a unified representation that enhances the understanding and processing of both visual and linguistic information.

As shown in Fig.~\ref{fig:placm}, $F_k$ and $F_W$  are concatenated on the sequence dimension to form $F_{k, W}$. Similarly, $F_k$ is concatenated as  to $F_S$  obtain $F_{k, S}$. To incorporate both vision and  linguistic information at the word level, we apply context modulation and language alignment on proposal points based on MHSA and MHCA. $F_q$ is first passed through an MHSA layer for initial self-encoding, and then through an MHCA layer to perform context modulation and word-level alignment together with $F_{k, W}$. Then, it is fed into a feed-forward network (FFN), including two fully-connected layers. The context modulation and word-level alignment is formulated as below:
\begin{align}
  F_{q|k,W} &=\operatorname{FFN}(\operatorname{MHCA}(\operatorname{MHSA}(F_q), F_{k, W}))
\end{align}

Similarly, we conduct context modulation and sentence-level alignment on $F_{q|k, W, S}$, formulated as:
\begin{align}
  F_{q|k, W, S} &=\operatorname{FFN}(\operatorname{MHCA}(\operatorname{MHSA}(F_{q|k,W}), F_{k, S}))
\end{align}

The processing of the PLACM module can be expressed as a mapping:
\begin{align}
F_q\rightarrow F_{q|k, W, S}.
\end{align}

We believe that this alignment mechanism enables the hierarchical capture of the relationships between each proposal point and individual words as well as the entire sentence, while the involvement of key points further provides latent informative relationships within the visual context. As shown in Fig.~\ref{fig:framework}, we emphasize the proposal points weighted by both word-level and sentence-level embeddings, which clearly exhibits the effectiveness of PLACM. Additional visualization results and analysis are presented in Section~\ref{sec:interesting}.

\subsection{Spatially Multi-Granular Modeling with PLACM }~\label{sec:SMGM}

The PLACM module offers exceptional flexibility, enabling it to accept pre-formulated proposals along with the context representations and the corresponding query text. This characteristic facilitates the pre-grouping of proposal points and key points, introduced in Section~\ref{sec:vision}, according to their spatial locations. As a result, to achieve a more comprehensive understanding of the scene and enhances the model's ability to ground objects across varying spatial scales, we design the spatially multi-granular modeling (SMGM) based on the PLACM module.

First, PLACM can be directly applied to the global field by incorporating the complete proposal points features $F_Q$ and key points features $F_K$:
\begin{align}
F_Q\rightarrow F_{Q|K, W, S}
\end{align}

The recent Swin-Transformer~\cite{swin-transformer} incorporates window-attention, introducing locality to the standard vision Transformer~\cite{ViT}. This approach highlights that 3D visual grounding narratives naturally consist of multiple local descriptions, in addition to a few global ones, to localize the target object progressively. For instance, the phrase ``There is a brown cabinet in the kitchen. It is on top of the refrigerator'' necessitates that the model first localizes the ``kitchen'' within the entire scene, and then localizes the ``brown cabinet'' within the local space, using the ``refrigerator'' as a reference. Consequently, we also conduct the PLACM on the local field.

As shown in Fig.~\ref{fig:framework}, we partition the entire scene into several cubic regions evenly, similar to Rubik's Cube. It yields a resolution of $r\times r \times r$. The proposal points and key points within the same region are grouped together and fed into the PLACM, expressed as:
\begin{align}
    F_{Q_r}\rightarrow F_{Q_r|K_r, W, S}
\end{align}
where $Q_r$ and $K_r$ represent the proposal points and key points within the same region, respectively.

Subsequently, $F_Q$ represents the global proposal features while $F_{Q_r}$ denotes a local one. These language-modulated proposal representation features are simply added to predict matching scores. The matching module, based on a straightforward MLP network, estimates the logits for each proposal bounding box. The bounding box with the highest score is selected as the final grounding box.

\subsection{Loss Function}~\label{sec:loss}
We follow the previous work~\cite{scanrefer,sat,ffl-3dog} with multi-task training, including 3D object detection, 3D visual grounding \textcolor{black}{and language auxiliary task}. Detection plays a critical role in the semantic representation of proposals. At the same time, we believe that this positive-intensive task will significantly benefit the grounding task with a severe imbalance between positives and negatives.

\textbf{Detection Loss.}
For the detection task, we adopt the same setting with VoteNet \cite{votenet} and GroupFree \cite{groupfree}, where the objectness prediction, box classification, center offset prediction, size classification, and size offset prediction are carried out for each proposal. Here we denote the object detection loss as $L_{D}$.

\textbf{Matching Loss.}
For the visual grounding, we follow ScanRefer \cite{scanrefer} to predict the matching scores $\{\hat{y_i}\}_{i=1}^M$ of $M$ proposal points. We set the label $y_i$ to one only if the box has the highest interaction-over-union (IoU) with the ground-truth box, and others are set to zeros. The ground-truth label is a one-hot encoding, so we apply softmax on matching scores. The cross-entropy loss is utilized as the matching loss function, formulated as:

\begin{align}
  L_{M} = -\sum_{i=1}^M y_i\text{log}(\hat{y_i})
\end{align}

\textbf{Language to object classification loss.}
The language description contains the target word, so an auxiliary task of language-to-object classification can be trained along with the visual grounding to classify the sentence-level feature $F_T$ to object classes, following ScanRefer~\cite{scanrefer}. Such classification regularizes the language model, supervised by the cross-entropy loss denoted as $L_{C}$. 

\textbf{Total Loss.}
The total loss $L_T$ is a weighted sum of the detection loss, matching loss, and language-to-object classification loss, denoted as:
\begin{align}
  L_T = \sum_{i} w_i * L_i, i \in \{D, M, C\}
\end{align}
where the $w_D$, $w_M$, and $w_C$ are hyper-parameters to indicate the importance of each task.

\subsection{Adaptation to Identification Paradigm}
\label{sec:adapting}

The aforementioned framework is designed for the grounding-by-detection paradigm as the benchmark method ScanRefer~\cite{scanrefer} defined. However, another benchmark dataset ReferIt3D~\cite{referit3d} introduces the problem of 3D visual grounding as an identification task, which provides the object ground truth boxes as known input for both the training and testing stage. Previous works~\cite{referit3d,instancerefer,beauty-detr,BUTD-DETR, multi-view-trans} usually group the point clouds of each provided instance at the beginning. However, such pre-processing fails to explore more contexts for point representation at the early stage.

Instead, we provide a simple but effective proposal aggregation mechanism for adapting to the identification setting. For the inputs with extra instance annotation, we retain the raw point clouds as inputs and the detection module. For all the proposal points at the last layer in the detection module, we group the proposal points by their instance annotations and adopt max-pooling for aggregating the instance-level features. After that, $M$ proposal points features are aggregated to $G$ instance features, where $G$ is the number of instances in the corresponding scene. For the subsequent PLACM, we use the instances features instead of the original proposal points features for finishing the modality fusion and prediction. 

We believe that this adaptation strategy is highly compatible with the grounding-by-detection training paradigm. Unlike the approaches~\cite{referit3d,instancerefer,beauty-detr,BUTD-DETR, multi-view-trans} of directly using ground truth to produce offline proposals, our approach preserves the learnable proposal generation throughout the grounding. Furthermore, it ensures the substantial utilization of contextual information from the detection phase during the point-language alignment.

\begin{table*}[t]

\begin{center}
\caption{Comparison with state-of-the-art methods on the ScanRefer~\cite{scanrefer} benchmark dataset.}
\begin{tabular}{llc|cccccccc}

\hline
\multirow{2}{*}{Method} & \multirow{2}{*}{Venue} & \multirow{2}{*}{Data} & \multicolumn{2}{c}{Unique} & \multicolumn{2}{c}{Multiple} & \multicolumn{2}{c}{Overall} \\
 & & & Acc@0.25 & Acc@0.5 & Acc@0.25 & Acc@0.5 & Acc@0.25 & Acc@0.5 \\
\hline 
 \multicolumn{2}{c}{ } & & & Validation Set \\
\hline
ScanRefer~\cite{scanrefer} & ECCV2020 & 3D & 67.64 & 46.19 & 32.06 & 21.26 & 38.97 & 26.10 \\ 
TGNN~\cite{TGNN} & AAAI2021 & 2D & 68.61 & 56.80 & 29.84 & 23.18 & 37.37 & 29.70 \\
InstanceRefer~\cite{instancerefer} & ICCV2021 & 3D & 77.45 & 66.83 & 31.27 & 24.77 & 40.23 & 32.93 \\
FFL-3DOG~\cite{ffl-3dog} & ICCV2021 & 3D & - & 67.94 & - & 25.70 & - & 34.01\\
3DVG-Transformer~\cite{3dvg-transformer} & ICCV2021 & 3D & 77.16 & 58.47 & 38.38 & 28.70 & 45.90 & 34.47 \\
3DJCG~\cite{3djcg} & CVPR2022 & 3D & 78.75 & 61.30 & 40.13 & 30.08 & 47.62 & 36.14\\
3D-SPS~\cite{3dsps} & CVPR2022 & 3D & 81.63 & 64.77 & 39.48 & 29.61 & 47.65 & 36.43 \\
BUTD-DETR~\cite{BUTD-DETR} & ECCV2022 & 3D & - & - & - & - & 52.2 & 39.8 \\

\hline
ScanRefer~\cite{scanrefer} & ECCV2020 & 2D + 3D & 76.33 & 53.51 & 32.73 & 21.11 & 41.19 & 27.40 \\
SAT~\cite{sat} & ICCV2021 & 2D + 3D & 73.21 & 50.83 & 37.64 & 25.16 & 44.54 & 30.14 \\
3DVG-Transformer~\cite{3dvg-transformer} & ICCV2021 & 2D + 3D & 81.93 & 60.64 & 39.30 & 28.42 & 47.57 & 34.67 \\ 

Multi-View Trans~\cite{multi-view-trans} & CVPR2022 & 2D + 3D & 77.67 & 66.45 & 31.92 & 25.26 & 40.80 & 33.26 \\ 
3D-SPS~\cite{3dsps} & CVPR2022 & 2D + 3D & \textbf{84.12} & 66.72 & 40.32 & 29.82 & 48.82 & 36.98 \\ 
3DJCG~\cite{3djcg} & CVPR2022 & 2D + 3D & 83.47 & 64.34 & 41.39 & 30.82 & \textbf{49.56} & 37.33 \\
D3Net~\cite{d3net} & ECCV2022 & 2D + 3D & - & \textbf{70.35} & - & 30.05 & - & 37.87 \\

\hline
\textbf{HAM} & & 3D & 79.24 & 67.86 & \textbf{41.46} & \textbf{34.03} & 48.79 & \textbf{40.60} \\

\hline 
 \multicolumn{2}{c}{ } & & & Online Benchmark \\
\hline

TGNN~\cite{TGNN} & AAAI2021 & 3D & 68.34 & 58.94 & 33.12 & 25.26 & 41.02 & 32.81 \\
InstanceRefer~\cite{instancerefer} & ICCV2021 & 3D & 77.82 & 66.69 & 34.57 & 26.88 & 44.27 & 35.80 \\
BUTD-DETR~\cite{BUTD-DETR} & ECCV2022 & 3D & \textbf{78.48} & 54.99 & 39.34 & 24.80 & 48.11 & 31.57 \\

\hline
ScanRefer~\cite{scanrefer} & ECCV2020 & 2D + 3D & 68.59 & 43.53 & 34.88 & 20.97 & 42.44 & 26.03 \\ 

3DVG-Transformer~\cite{3dvg-transformer} & ICCV2021 & 2D + 3D & 77.33 & 57.87 & 43.70 & 31.02 & 51.24 & 37.04 \\
 
3DJCG~\cite{3djcg} & CVPR2022 & 2D + 3D & 76.75 & 60.59 & \textbf{43.89} & 31.17 & \textbf{51.26} & 37.76 \\

D3Net~\cite{d3net} & ECCV2022 & 2D + 3D & - & \textbf{68.43} & - & 30.74 & - & 39.19 \\

\hline
\textbf{HAM}  & & 3D & 77.99 & 63.73 & 41.48 & \textbf{33.24} & 49.67 & \textbf{40.07} \\
\hline
\end{tabular}
\label{table:scanrefer}
\end{center} 
\end{table*}

\section{Experiments}
\label{experiments}
\subsection{Implementation Details}

For the point cloud encoding, the number of the backbone input points $L$ is set to 50,000. The number of the key points $N$ is set to 1,024, and the number of the proposal points $M$ is set to 512. All of the feature dimension $C$ is set to 288. The number of the multi-head cross-attention layers in the proposal generation module is set to 12. In Section~\ref{sec_ablation}, we also provide and discuss a lighter-weight point cloud backbone in the ablation studies. For the proposal aggregation, the maximum instance number $G$ is set to 88. For the \textcolor{black}{intra-sentence ensemble}, we randomly sample 1 to 6 sentences corresponding to one scene for fusing into the composited sentence. And for the \textcolor{black}{inter-sentence ensemble}, we attach eight sentences (randomly sampled from the composited and original sentences) to one scene for dense grounding. The maximum sentence length $T$ is set to 200.
 
For the positional embedding of all the self-attention and cross-attention layers, following Group-Free \cite{groupfree}, the 3D coordinates of the key points are mapped to a positional embedding of the dimension $C$ via a fully-connected layer and then added to the \textcolor{black}{corresponding features}. Similarly, the central location and size of the proposal bounding boxes are mapped and added to the proposal points features. For the \textcolor{black}{word-level embedding and sentence-level embedding}, we utilize the regular sine-cosine functions as position embedding.

For the SMGM, we provide an efficient implementation based on mask attention. Notably, we retain the attention weights of the points in the same region only, while others are masked out by filling with infinity before the Softmax operation in the cross-attention layer. Such masked attention can be seamlessly aggregated into PLACM module with high parallelism easily. The default space partition resolution $r$ is set to $4$.

Besides, we adopt the AdamW optimizer\cite{adamw} and set the initial learning rate to $2 \times 10^{- 4}$ for the attention layers and $1 \times 10^{- 2}$ for others. The loss weights $w_D$, $w_M$, and $w_C$ are 10, 0.1, and 0.1. We train the network by 200 epochs, with a batch size of 4. Codes are implemented by Pytorch and run on 8 Tesla-V100 GPUs.

\begin{table*}[t]
\setlength\tabcolsep{3pt}
\begin{center}
\caption{Comparison with state-of-the-art methods on the ReferIt3D~\cite{referit3d} benchmark dataset.}
\begin{tabular}{clcccc|ccccc}

\hline
 Subset & Method & Venue & Pre-training & Text Encoder & Data   & Easy & Hard & View-dep. & View-indep. & Overall\\
\hline 

\hline
\multirow{11}{*}{Nr3D} & ReferIt3D~\cite{referit3d} & ECCV2020 & \XSolidBrush & RNN~\cite{lstm}  & 3D & 43.6 & 27.9 & 32.5 & 37.1 & 35.6 \\
 & TGNN~\cite{TGNN} & AAAI2021 & \XSolidBrush & BERT~\cite{devlin2018bert}  & 3D & 44.2 & 30.6 & 35.8 & 38.0 & 37.3 \\
 & InstanceRefer~\cite{instancerefer} & ICCV2021 & \Checkmark & GRU~\cite{gru}  & 3D & 46.0 & 31.8 & 34.5 & 41.9 & 38.8 \\
 & 3DVG-Transformer~\cite{3dvg-transformer} & ICCV2021 & \XSolidBrush & GRU~\cite{gru} & 3D & 48.5 & 34.8 & 34.8 & 43.7 & 40.8 \\
 & FFL-3DOG~\cite{ffl-3dog} & ICCV2021 & \XSolidBrush & GRU~\cite{gru} & 3D  & 48.2 & 35.0 & 37.1 & 44.7 & 41.7 \\
 & TransRefer3D~\cite{transrefer3d} & ACM MM2021 & \XSolidBrush & GRU~\cite{gru} & 3D & 48.5 & 36.0 & 36.5 & 44.9 & 42.1  \\
 & LanguageRefer~\cite{languagerefer} & CoRL2021 & \XSolidBrush & DistilBert~\cite{distilbert} & 3D  & 51.0 & 36.6 & 41.7 & 45.0 & 43.9 \\
 & SAT~\cite{sat} & ICCV2021 & \XSolidBrush & BERT~\cite{devlin2018bert} & 2D + 3D & 56.3 & 42.4 & 46.9 & 50.4 & 49.2  \\

 & Multi-View Trans~\cite{multi-view-trans} & CVPR2022 & \Checkmark & BERT~\cite{devlin2018bert} & 2D + 3D  & 61.3 & 49.1 & \textbf{54.3} & 55.4 & 55.1 \\
 & 3D-SPS~\cite{3dsps} & CVPR2022 & \Checkmark & CLIP~\cite{clip} & 2D + 3D  & 58.1 & 45.1 & 48.0 & 53.2  & 51.5\\
 & BUTD-DETR~\cite{BUTD-DETR} & ECCV2022 & \Checkmark & RoBERTa~\cite{roberta} & 3D  & 60.7 & 48.4 & 46.0 & 58.0 & 54.6  \\
\hline
 & \textbf{HAM}  &  & \XSolidBrush & GRU~\cite{gru} & 3D   & 54.3 & 41.9 & 41.5 & 51.4 & 48.2\\
 & \textbf{HAM+BUTD-DETR}  &  & \Checkmark & RoBERTa~\cite{roberta} & 3D  & \textbf{64.7} & \textbf{49.7} & 49.0 & \textbf{60.5} & \textbf{57.2} \\
\hline

\hline 
 \multirow{10}{*}{Sr3D} & ReferIt3D~\cite{referit3d} & ECCV2020 & \XSolidBrush & RNN~\cite{lstm} & 3D  & 44.7 & 31.5 & 39.2 & 40.8 & 40.8 \\
 & TGNN~\cite{TGNN} & AAAI2021 & \XSolidBrush & BERT~\cite{devlin2018bert} & 3D 
  & 48.5 & 36.9 & 45.8 & 45.0 & 45.0\\
 & InstanceRefer~\cite{instancerefer} & ICCV2021 & \Checkmark & GRU~\cite{gru} & 3D   & 51.1 & 40.5 & 45.4 & 48.1 & 48.0\\
 & 3DVG-Transformer~\cite{3dvg-transformer} & ICCV2021 & \XSolidBrush & GRU~\cite{gru} & 3D  & 54.2 & 44.9 & 44.6 & 51.7 & 51.4  \\
 & LanguageRefer~\cite{languagerefer} & CoRL2021 & \XSolidBrush & DistilBert~\cite{distilbert} & 3D  & 58.9 & 49.3 & 49.2 & 56.3 & 56.0 \\
 & TransRefer3D~\cite{transrefer3d} & ACM MM2021 & \XSolidBrush & GRU~\cite{gru} & 3D  & 60.5 & 50.2 & 49.9 & 57.7  & 57.4\\
 & SAT~\cite{sat} & ICCV2021 & \XSolidBrush & BERT~\cite{devlin2018bert} & 2D + 3D  & 61.2 & 50.0 & 49.2 & 58.3 & 57.9 \\
 & Multi-View Trans~\cite{multi-view-trans} & CVPR2022 & \Checkmark & BERT~\cite{devlin2018bert} & 2D + 3D   & 66.9 & 58.8 & \textbf{58.4} & 64.7 & 64.5\\
 & 3D-SPS~\cite{3dsps} & CVPR2022 & \Checkmark & CLIP~\cite{clip} & 2D + 3D & 56.2 & \textbf{65.4} & 49.2 & 63.2  & 62.6 \\
 & BUTD-DETR~\cite{BUTD-DETR} & ECCV2022 & \Checkmark & RoBERTa~\cite{roberta} & 3D & 68.6 & 63.2 & 53.0 & 67.6 & 67.0  \\
\hline
 & \textbf{HAM}  &  & \XSolidBrush & GRU~\cite{gru} & 3D  & 65.9 & 54.6 & 52.5 & 63.0 & 62.5\\ 
 & \textbf{HAM+BUTD-DETR}  &  & \Checkmark & RoBERTa~\cite{roberta} & 3D  & \textbf{71.3} & 63.6 & 54.1 & \textbf{69.6} & \textbf{69.0}\\ 

\hline
\end{tabular}
\label{table:referit3d}
\end{center} 
\end{table*}

\begin{table}[t]
    \caption{Comparison with state-of-the-art methods on the ReferIt3D Nr3D subset by Acc@0.25 and Acc@0.5. All the  results of other methods are provided by the D3Net~\cite{d3net}.}
    \label{table:referit3d_a25a50}
    \centering
    \begin{tabular}{l|ccc}
    \hline
        Methods & Unique & {Multiple} & {Overall} \\

    \hline
         &      & Acc@0.5 & \\
    \hline
    ScanRefer~\cite{scanrefer} & - & 12.17 & 12.17 \\
    3DVG-Transformer~\cite{3dvg-transformer} & - & 14.22 & 14.22 \\
    D3Net~\cite{d3net} & - & 25.23 & 25.23 \\ 
    \hline
    \textbf{HAM} & -  & \textbf{27.11} & \textbf{27.11} \\ 

    \hline
    
    \end{tabular}
    
\end{table}

\subsection{Dataset and Evaluation Metric}

ScanRefer \cite{scanrefer} and ReferIt3D \cite{referit3d} are two benchmark datasets released recently for visual grounding on point clouds, which are both conducted on the ScanNet~\cite{scannet} dataset but take different evaluation metrics.

\textbf{ScanRefer.}
It follows the grounding-by-detection paradigm as the 2D visual grounding on images that only raw point clouds and query sentences are given. The ScanRefer dataset \cite{scanrefer} augments the 3D instance-level indoor dataset ScanNet \cite{scannet} with 51,583 human-written free-form descriptions of 11,046 objects in 3D scans. It splits the training/validation/testing set with 36,655, 9,508, and 5,410 samples, respectively. We follow the evaluation metric \emph{Acc@IoU=0.25} and \emph{Acc@IoU=0.5} proposed by ScanRefer to evaluate all the methods. Specifically, they measure the percentage of the positively predicted bounding boxes whose IoUs with the ground-truth bounding boxes are higher than 0.25 and 0.5, respectively. Additionally, scenes with a sole target object without similar duplicates are labeled as ``Unique'' and others as ``Multiple'' alternatively. Both ``Unique'' and ``Multiple'' are unified as ``Overall''. We report all these evaluation metrics on the validation set and online testing set.

\textbf{ReferIt3D.} 
It contains two sub-datasets, where a sub-dataset is with natural-language reference annotations (\textit{Nr3D}), and another sub-dataset is with synthetic reference annotations (\textit{Sr3D}). \textit{Nr3D} contains 41,503 samples collected by ReferItGame~\cite{referitgame}, an online reference labeling game on 3D scenes. \textit{Sr3D} contains 83,572 samples collected by an automatic template-based generator. Different from ScanRefer, ReferId3D casts visual grounding to a \textcolor{black}{identification task} and assumes the \textcolor{black}{object ground truth boxes} of objects is known during training and inference. During the evaluation, ReferIt3D only needs to calculate the identification accuracy. Similar to ScanRefer, \textit{Nr3D} and \textit{Sr3D} also are further split into different subsets by the sample attributes, where the ``easy'' and ``hard'' subsets are defined the same as ``Unique'' and ``Multiple'' in ScanRefer, and the ``view-dep'' and ``view-indep'' subsets are defined by whether the referring queries are dependent or independent on the camera view. We propose a proposal aggregation strategy for adapting our grounding-by-detection framework to such identification paradigm with the given instance annotations, which is fully introduced in Section \ref{sec:adapting}, and we follow the evaluation metric \emph{Accuracy} proposed by the ReferIt3D~\cite{referit3d} to evaluate our method. We also provide \emph{Acc@IoU=0.5} results on ReferIt3D by removing the proposal aggregation following D3Net~\cite{d3net}. \textcolor{black}{Since there is no ``Unique'' case in ReferIt3D, we report results on ``Multiple'' and ``Overall'' subsets as D3Net does.}

\begin{figure*}[t]
    \centering
    \includegraphics[height=9.cm]{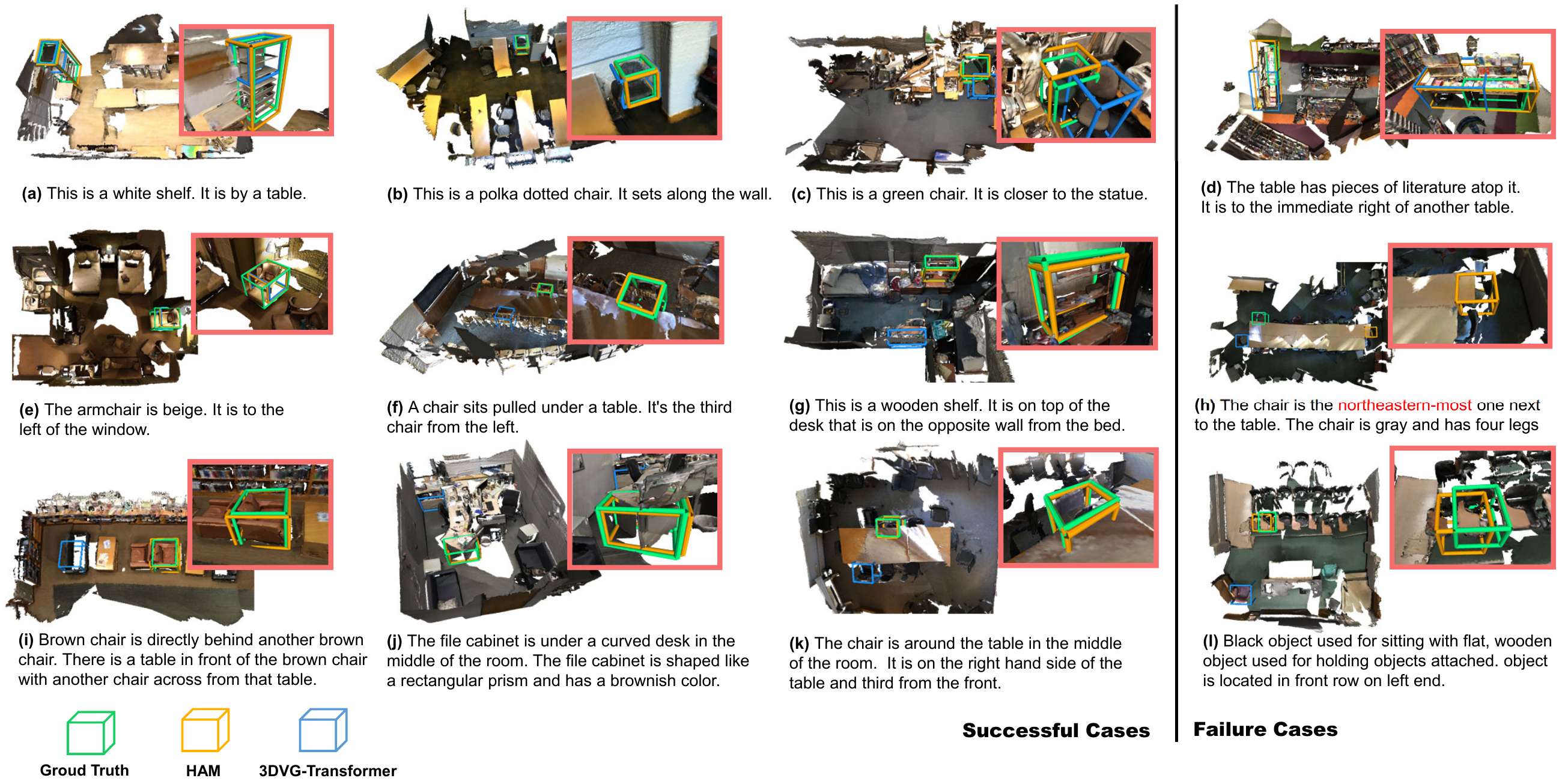}
    \caption{The visualization results of the proposed method.}
    \label{fig:vis}
\end{figure*}

\subsection{Performance Comparisons with the State-of-the-art Methods}

\textbf{Quantative Results.} In Table~\ref{table:scanrefer} and Table~\ref{table:referit3d}, we compare our HAM with the state-of-the-art methods~\cite{scanrefer,TGNN,instancerefer,ffl-3dog,3dvg-transformer,3djcg,3dsps,BUTD-DETR,languagerefer,multi-view-trans,languagerefer,sat,d3net} on both ScanRefer~\cite{scanrefer} and ReferIt3D~\cite{referit3d} Nr3D/Sr3D benchmark datasets. For a fair comparison, we report the results according to the modalities used as input data. Here ``3D'' means only the raw attributes such as coordinates, color, and normal vectors of the raw point clouds are used. ``2D + 3D'' means using the 2D multi-view features as extra inputs.

Table~\ref{table:scanrefer} demonstrates the quantitative results on both the validation set and the online benchmark of the ScanRefer~\cite{scanrefer}.  Here HAM only takes the ``3D'' inputs with efficient end-to-end training, and it outperforms all the state-of-the-art methods on the validation set and online benchmark. It is also the champion on the ECCV 2022 ScanRefer Challenge with the highest overall Acc@0.5. \textcolor{black}{For the most challenging metric  ``Multiple'' that expects the grounder to locate the target among multiple objects with the same category, HAM ranks first among all the methods in the validation set for both ``Acc@0.25'' and ``Acc@0.5'', and it also ranks first in the online benchmark for ``Multiple Acc@0.5''}, which shows its remarkable capacity for 3D visual grounding in complex environments. In addition, without any extra 2D multi-view features and pre-training for the object-detector utilized in a bulk of works \cite{d3net,BUTD-DETR,instancerefer}, our method still has a promising result on ``Unique Acc@0.5'' compared to D3Net~\cite{d3net}. 

Table~\ref{table:referit3d} lists the comparative results on the ReferIt3D~\cite{referit3d} Nr3D and Sr3D datasets. Even without leveraging a pre-training backbone network, object detector, or advanced text encoder, HAM outperforms all competing methods under identical settings. Moreover, HAM delivers results comparable to state-of-the-art techniques. For instance, HAM achieves 48.2\% accuracy versus SAT's 49.2\% on Nr3D and 62.5\% versus 3D-SPS's 62.6\% on Sr3D. Such results highlight the effective integration of localization and identification within our framework.

Additionally, to further validate HAM's capability with more advanced text encoders and pre-training, we incorporate the text encoding and the proposal encoding methods used in BUTD-DETR~\cite{BUTD-DETR}, denoted as ``HAM+BUTD-DETR''. Specifically, following BUTD-DETR, all the ground truth boxes are extracted offline and mapped to visual tokens for enhancing proposals in the encoder, where the point cloud backbone network and the text encoder RoBERTa~\cite{roberta} are pre-trained. And we implement PLACM and SMGM upon such encoders in these experiments. We recommend readers to refer to BUTD-DETR for more technical details. The experimental results listed in Table~\ref{table:referit3d} demonstrate that our method outperforms the  state-of-the-art methods in both the Nr3D and Sr3D subsets. For example, ``HAM+BUTD-DETR'' remarkably surpasses Multi-View Trans by 2.1\% on Nr3D and outperforms BUTD-DETR by 2.0\% on Sr3D.

We also report Acc@0.5 on the ReferIt3D in Table~\ref{table:referit3d_a25a50}. Surprisingly, HAM achieves the best performance again by a significant margin compared to D3Net, ranking second on the more challenging ScanRefer. In summary, we claim that HAM deeply fuses the grounding and detection compared to any previous work that treats detection as a plug-in.

\noindent\textbf{Qualitative Results}. We visualize several success cases and failure cases of our proposed method as shown in Fig.~\ref{fig:vis}. For comparison, we also visualize the predictions of 3DVG-Transformer~\cite{3dvg-transformer} simultaneously. The results from the first row to the third rows show that HAM well understands texts of various lengths, from short to long, thanks to the proposed word-level and sentence-level embeddings. Most of the successful cases correctly understand clues from reference objects in texts, for example, ``by a table'' in Fig.~\ref{fig:vis} (a), ``closer to the statue'' in Fig.~\ref{fig:vis} (c), and more complex cases such as ``on the right hand side of the table and third from the front'' in Fig.~\ref{fig:vis} (k). For the ``Multiple'' samples, which require grounding the target among multiple objects with the same class, such as Fig.~\ref{fig:vis} (c)(e)(f)(i)(k), HAM demonstrates much more excellent discrimination ability, compared to 3DVG-Transformer. 

For the failure cases, we summarize three main reasons. The first is the mismatch between the text and the scene. For example, in Fig.~\ref{fig:vis} (d), it is difficult to distinguish two tables in the raw point cloud, even for humans. And HAM mistakenly identifies two closely spaced tables as one. The second is the ambiguity from the perspective-related description. As shown in Fig.~\ref{fig:vis} \textcolor{black}{(h)}, the word ``northeastern-most'' is ambiguous when the perspective is not clear, and HAM predicts the chair on the opposite side of the ground truth box. The third is the incomplete point clouds due to scanning and reconstruction. As shown in Fig.~\ref{fig:vis} (I), the points of the target object are incomplete, and it further causes the inaccurate localization of HAM. We believe
that alleviating language ambiguity and conducting point cloud completion can significantly contribute to this challenging task, where we leave it in the future.

\subsection{Ablation Studies}\label{sec_ablation}

In this section, we first conduct an ablation study on ScanRefer~\cite{scanrefer} to verify the effectiveness of each component in the proposed HAM. We then carefully ablate the setting of each key component on ScanRefer. 

\noindent\textbf{Ablation study on each proposed component.} We conduct ablation experiments to evaluate each proposed component, as detailed in Table~\ref{table:ablation}. To ease the ablation, we adopt a lighter-weight point cloud backbone (\textbf{LB}) instead of the default backbone (\textbf{DB}), which consists $l=6$ (12 by defaults) decoder layers and $M=256$ (512 by defaults) proposal points. Additionally, all experiments equip the word masking and the concentration sampling strategies.

We first conduct a baseline setting called ``\textbf{PL-Concat}'', which expanded on ScanRefer~\cite{scanrefer}, by replacing its point cloud and language encodings with the same ones used in HAM and concatenating point-language features in the same manner as ScanRefer, to ensure a fair comparison. To demonstrate the effectiveness of our PLACM, we directly compared it with PL-Concat.
Next, we conduct ablation experiments on the PLACM  and SMGM, which are equipped with two prompt engineering techniques: inter-sentence ensemble (\textbf{Inter-SE}) and intra-sentence ensemble (\textbf{Intra-SE}), to comprehensively and gradually evaluate their effectiveness.

From the experimental results, we find that incrementally stacking each of the proposed modules or strategies of the whole framework brings performance gains gradually on both the Acc@0.25 and the Acc@0.5.

\begin{table}[t]
\setlength\tabcolsep{2pt}
\renewcommand\arraystretch{1.02}
    \centering
    \caption{The ablations on each proposed component.}
    \begin{tabular}{l|cc}
	\hline
	Network Setting & {Acc@0.25} & {Acc@0.5} \\
	\hline 
        LB + PL-Concat & 38.87 & 31.69 \\
	LB + PLACM & 43.76 & 34.01 \\ 
    \hline 
	LB + PLACM + Inter-SE & 45.30 & 36.47 \\
        LB + PLACM + Inter-SE + Intra-SE & 47.18 & 38.45 \\
	DB + PLACM + Inter-SE + Intra-SE & 47.50 & 40.22 \\ 
        \hline 
        LB + PLACM + SMGM + Inter-SE  & 46.42  & 38.15 \\
         LB + PLACM + SMGM + Inter-SE + Intra-SE & 47.65  & 38.86 \\
	DB + PLACM + SMGM + Inter-SE + Intra-SE  & \textbf{48.79} & \textbf{40.60} \\ 
	
	\hline
    \end{tabular}
    \label{table:ablation}
\end{table}

\begin{table}[t]
\setlength\tabcolsep{10pt}
    \centering
    \caption{The ablations on down-sampling strategies (under "LB + PLACM" setting).}
    \begin{tabular}{c|cc}
	\hline
	Sampling Strategy & Acc@0.25 & Acc@0.5\\
	\hline
	D-FPS & 43.72 & 32.92\\
	F-FPS & 43.24 & 33.46\\
	FS & 42.76 & 32.40\\
	 \textbf{CS} & \textbf{43.76} & \textbf{34.01}\\
	\hline
	\end{tabular}
    \label{table:fps}
\end{table}

\vspace{0.1cm}
\noindent\textbf{The Concentration Sampling Strategy.} To explore the effective and efficient sampling strategy for the raw input point clouds with additional features such as colors and normal vectors, we compare our concentration sampling (CS) strategy with the regular distance-FPS (D-FPS), feature-FPS (F-FPS) and the fusion sampling (FS)~\cite{3dssd} on the baseline model. The corresponding results are shown under "LB + PLACM" setting in Table~\ref{table:ablation}.

\begin{table}[t]
    \setlength\tabcolsep{2pt}
    \centering
    \caption{The ablations on language augmentation (under ``LB + PLACM + Inter-SE + Intra-SE'' setting).}
    \begin{tabular}{cc|cc}
    \hline
    \textcolor{black}{\#Sentences for Inter-SM} & \textcolor{black}{\#Sentences for Intra-SM} & Acc@0.25 & Acc@0.5\\
    \hline
    32 & 1 & 42.76 & 32.40 \\
    16 & 1 & 45.39 & 35.58 \\
    8 & 1 & \textbf{45.30} & \textbf{36.47} \\
    1 & 1 & 41.20 & 32.24 \\
    \hline
    8 & [1, 3] & 46.63 & 38.10 \\
    8 & [1, 6] & \textbf{47.18} & \textbf{38.45} \\
    \hline
    \end{tabular}
    \label{table:textual-aug}
\end{table}

\begin{table}[t]
    \setlength\tabcolsep{10pt}
    \centering
    \caption{The ablations on different partition resolutions for SMGM (under "LB + PLACM + SMGM + Inter-SE" setting).}
    \begin{tabular}{c|cc}
	\hline
	Partition Resolution & Acc@0.25 & Acc@0.5 \\
	\hline
	without SMGM & 45.30 & 36.47 \\
	$3 \times 3 \times 3$ & 47.08 & 37.87 \\
	$4 \times 4 \times 4$ & \textbf{46.42} & \textbf{38.15} \\
	$5 \times 5 \times 5$ & 46.15 & 36.35 \\
	\hline
	\end{tabular}
    \label{table:sla}
\end{table}

As shown in Table~\ref{table:fps}, taking the colors and normal vectors of the points into account of the down-sampling, the F-FPS gains 0.5\% improvement on the Acc@0.5, which means that such additional attribute inputs help filter out more discriminative and meaningful points. However, when utilizing the fusion sampling, the performance drops significantly. We believe that a large amount of overlap (around 20\%) between the two sampling strategies causes the shortage of high-quality points for the backbone network and the proposal generation. In our concentration sampling setting, the points filtered out by the D-FPS and F-FPS are even without overlapping. Such a simple yet effective strategy achieves the best performance than other down-sampling strategies on both Acc@0.25 and Acc@0.5. 

\vspace{0.1cm}
\noindent\textbf{\textcolor{black}{The Strategies of Prompt Engineering.}} To further explore the suitable settings for \textcolor{black}{prompt engineering, we conduct a series of experiments for both inter-sentence and intra-sentence ensembles.} To ease the validation, we adopt the ``LB + PLACM + Inter-SE + Intra-SE'' network setting in Table~\ref{table:ablation}. 

For the inter-sentence ensemble, which attaches multiple sentences to a single point cloud scene, ablations for the number of sentences for attaching are conducted. As shown in Table~\ref{table:textual-aug}, without the intra-sentence ensemble, the ``8 Sentences for Inter-SM'' setting performs better than more or fewer sentences. A potential reason is that our framework is trained under the end-to-end grounding-by-detection paradigm. Thus the inter-sentence ensemble needs to make a trade-off between visual grounding and object detection. When more sentences are associated with a single scene in one training sample batch, the grounding becomes more intensive. In summary, a proper number of sentences for the inter-sentence ensemble is important for those grounding-by-detection methods including HAM.

For the intra-sentence ensemble, it provides another enhancement of the raw sentences. We provide two options in Table~\ref{table:textual-aug}, where $[1, 3]$ and $[1, 6]$ mean that we augment the sentences by a random number from 1 to 3 or from 1 to 6, respectively. The result shows that combining more sentences into a single sentence brings more performance gain. 

\begin{table}[t]
    \centering
    \caption{The ablations on different feature inputs (under "DB + PLACM + SMGM + Inter-SE + Intra-SE" setting). }
	\begin{tabular}{l|cc}
	\hline
	Feature(s) & Acc@0.25 & Acc@0.5\\
	\hline
        xyz & 47.36 & 38.91 \\
	xyz + rgb & 47.75 & 39.91 \\
	xyz + normals & 47.50 & 39.46 \\
	xyz + multi-view & 47.77 & 37.69 \\
	xyz + rgb + normals & \textbf{48.79}  & \textbf{40.60} \\
        xyz + rgb + normals + multi-view & 46.43 & 36.25 \\
	\hline
	\end{tabular}
    \label{table:features}
\end{table}

\begin{table}[t]
    \centering
    \caption{The ablations on different adapting strategies on Nr3D dataset (under ``LB+PLACM'' setting).}
	\begin{tabular}{l|ccc}
	\hline
	Adapting Strategy & Acc@0.25 & Acc@0.5 & Overall Acc \\
	\hline
	Center Distance & 37.63 & 27.11 & 37.90 \\
	IoU & 37.63 & 27.11 & 38.34 \\
	Mean-pooling & - & -  & 42.40 \\
	Max-pooling & - & - & \textbf{43.62} \\
	\hline
	\end{tabular}
    \label{table:adapting}
\end{table}

\noindent\textbf{The Partition Setting of the SMGM}. We further conduct a series of experiments for exploring the proper partition setting of the SMGM. To ease the validation, we adopt the ``LB + PLACM + SMGM + Inter-SE'' network setting in Table~\ref{table:ablation}. The experimental results are shown in Table~\ref{table:sla}.  We find that compared with the baseline without introducing SMGM, both $3 \times 3 \times 3$ and $4 \times 4 \times 4$ resolutions achieve better performance. However, when adopting the higher partition resolution $5 \times 5 \times 5$, the performance is worse. We assume that though SMGM introduces constrained local attention, the number of key points and proposal points in a local region will decrease as the resolution increases. Thus, we use the resolution $4 \times 4 \times 4$ for the default setting in HAM.

\noindent\textbf{Different Point Cloud Features}. To explore the impact of different point cloud features (\ie~colors, normals, and multi-view features) on the performance of our model, we conduct an ablation study on ScanRefer~\cite{scanrefer}, where we train and evaluate the model using different combinations of such features. We use the multi-view features extracted from ENet~\cite{enet} in line with ScanRefer. To ease the validation, we adopt the ``DB + PLACM + SMGM + Inter-SE + Intra-SE'' network setting in Table~\ref{table:ablation}.  Table \ref{table:features} shows the results. Using only the coordinates as input results in an Acc@0.5 of 38.91, which surpasses the performance of the majority of state-of-the-art methods. Adding colors or normals to the coordinates can yield slight improvements in accuracy. The combination of coordinates, colors, and normals results in the best performance. However, the addition of multi-view features leads to a decrease in performance, suggesting that multi-view features may not contribute significantly to our model. We hypothesize that introducing high-dimensional multi-view features adds complexity to the model and may lead to overfitting. Furthermore, utilizing multi-view features that rely on a pre-trained extractor is inconsistent with our motivation for an end-to-end framework. Hence, we did not include it in our default setting.

\noindent\textbf{Different Strategies for Adapting to Identification Paradigm}. We conduct a bulk of experiments on adapting our grounding-by-detection HAM framework to the identification paradigm for the ReferIt3D~\cite{referit3d} dataset, which provides the prior object ground truth boxes for all objects. To ease the validation, we adopt the ``LB + PLACM'' network setting in Table~\ref{table:ablation}. The detection branch taking the raw point clouds as inputs is retained for providing high-quality key points features for the following point-language attention. A simple yet effective strategy is to assign the predicted bounding box to the object ground truth box with a minimal center distance. Another similar solution is to use the object ground truth boxes that have the maximum intersection over union (IoU) with the predicted box as a result. These two strategies only 
use the  object ground truth boxes in the prediction stage without changing any network structure and intermediate features. As shown in Table~\ref{table:adapting}, these two simple strategies achieve comparable performance with some state-of-the-art methods such as TGNN~\cite{TGNN} and InstanceRefer~\cite{instancerefer}.

In addition, we try to aggregate all the proposal points within each  object ground truth box by mean-pooling or max-pooling on features, as aforementioned in Section~\ref{sec:adapting}. Then the proposal-level features are converted into instance-level features. We observe a significant performance improvement in Table~\ref{table:adapting}, compared to the assignment strategies with center distance and IoU. We claim that the aggregation from multiple meaningful proposal features acts as a voting mechanism.

\begin{figure}[t]
    \centering
    \includegraphics[height=5cm]{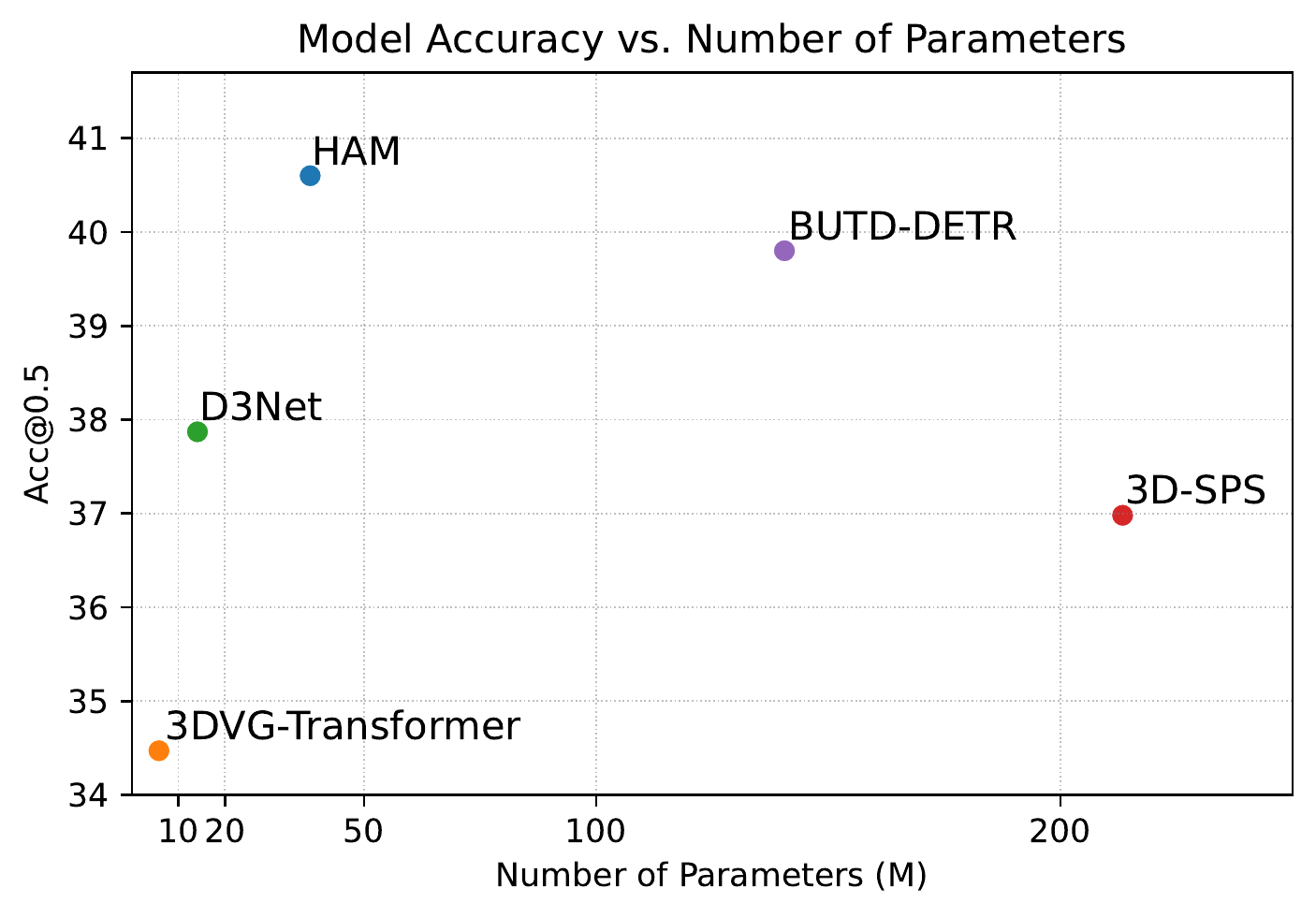}
    \caption{Model Accuracy vs. Number of Parameters.}
    \label{fig:acc_vs_param}
\end{figure}

\begin{figure}[t]
    \centering
    \includegraphics[height=5cm]{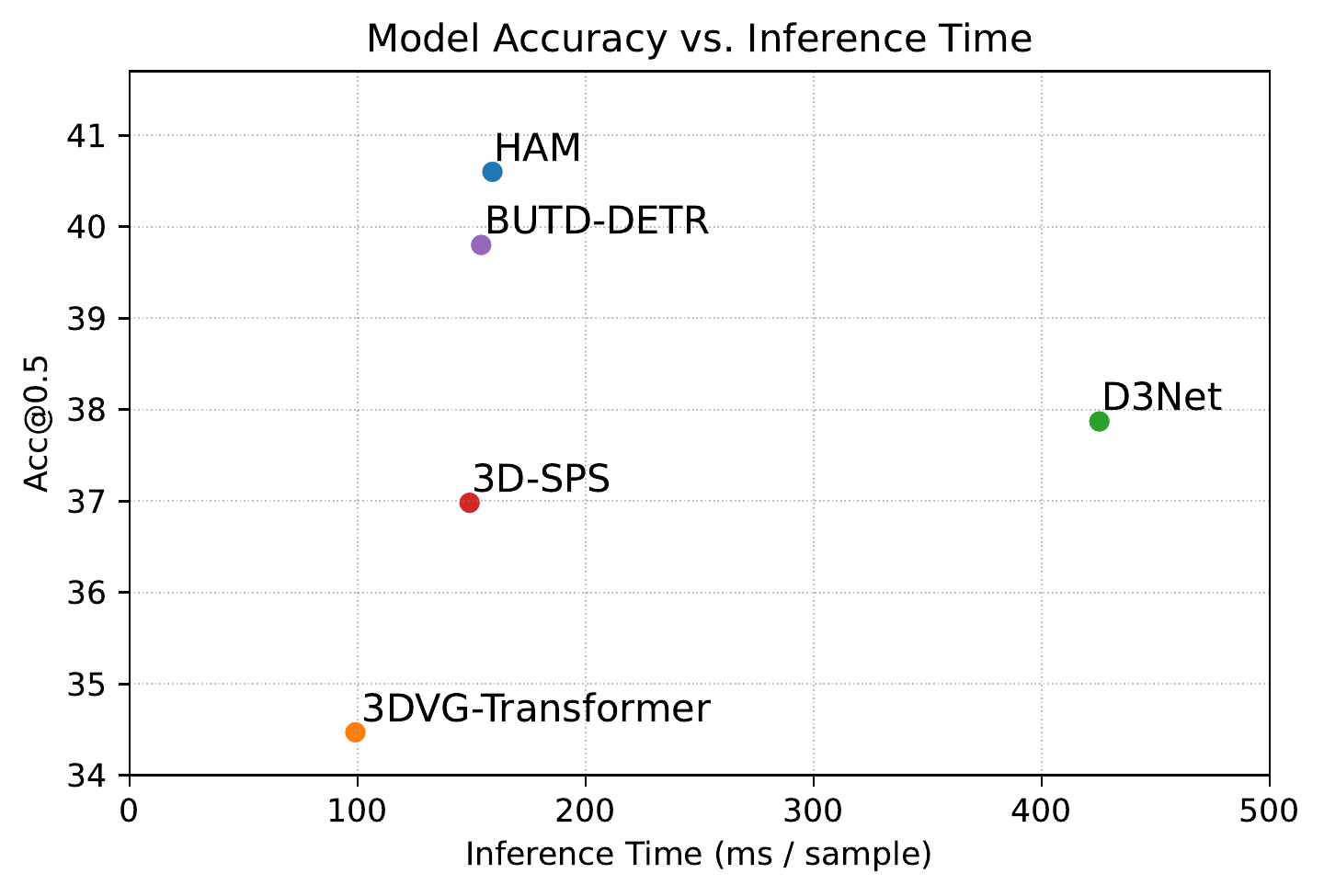}
    \caption{Model Accuracy vs. Inference Time.}
    \label{fig:acc_vs_infer}
\end{figure}

\begin{figure}[t]
	\centering
	\includegraphics[height=7.0cm]{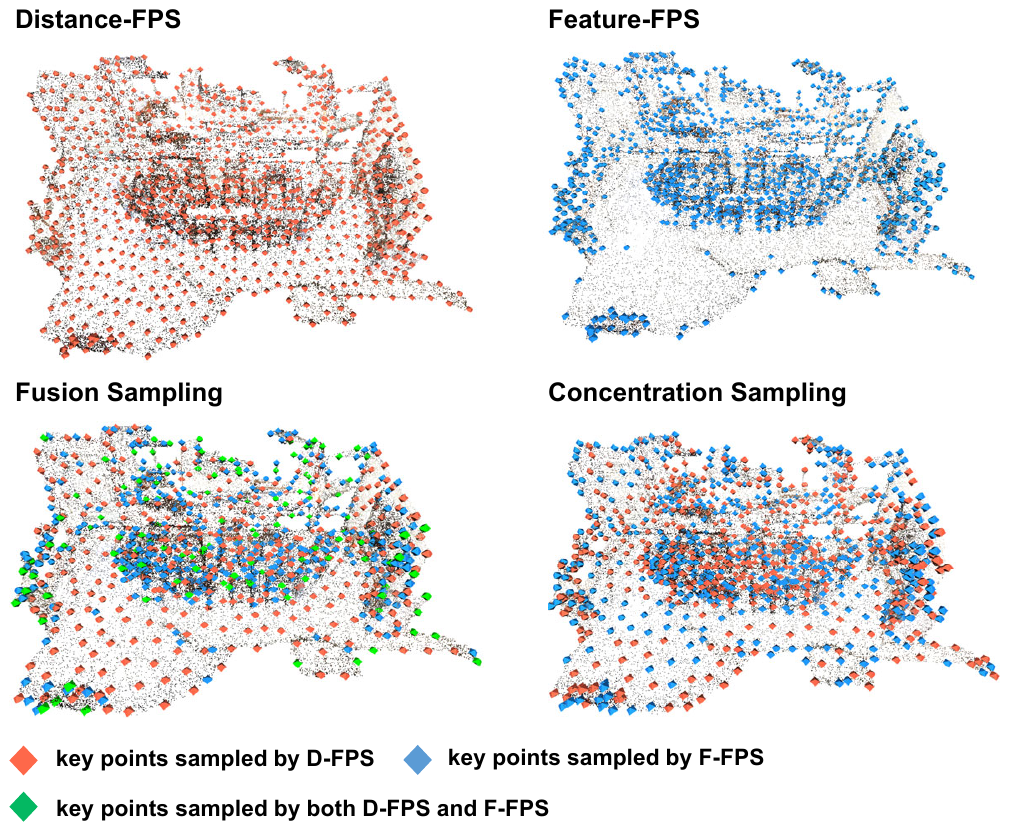}
	
	\caption{The visualization of different point sampling strategies. Sampling the same number of points, concentration sampling not only ensures the coverage of the entire space but also extracts a sufficient number of high-quality foreground points.}
	\label{fig:fps}
\end{figure}

\noindent\textbf{Efficiency comparison}. To further validate the efficiency of HAM, we conducted experiments on ScanRefer~\cite{scanrefer} to evaluate the inference time with reference to the number of parameters. As shown in Figs.~\ref{fig:acc_vs_param} and \ref{fig:acc_vs_infer}, 
the number of parameters and the inference speed of HAM are at a moderate level compared to other methods. However, HAM achieves the highest Acc@0.5, indicating that it has achieved a promising balance between efficiency and performance. We shall also explore more efficient models in the future work.

\begin{figure}[t]
    \centering
    \includegraphics[height=2.8cm]{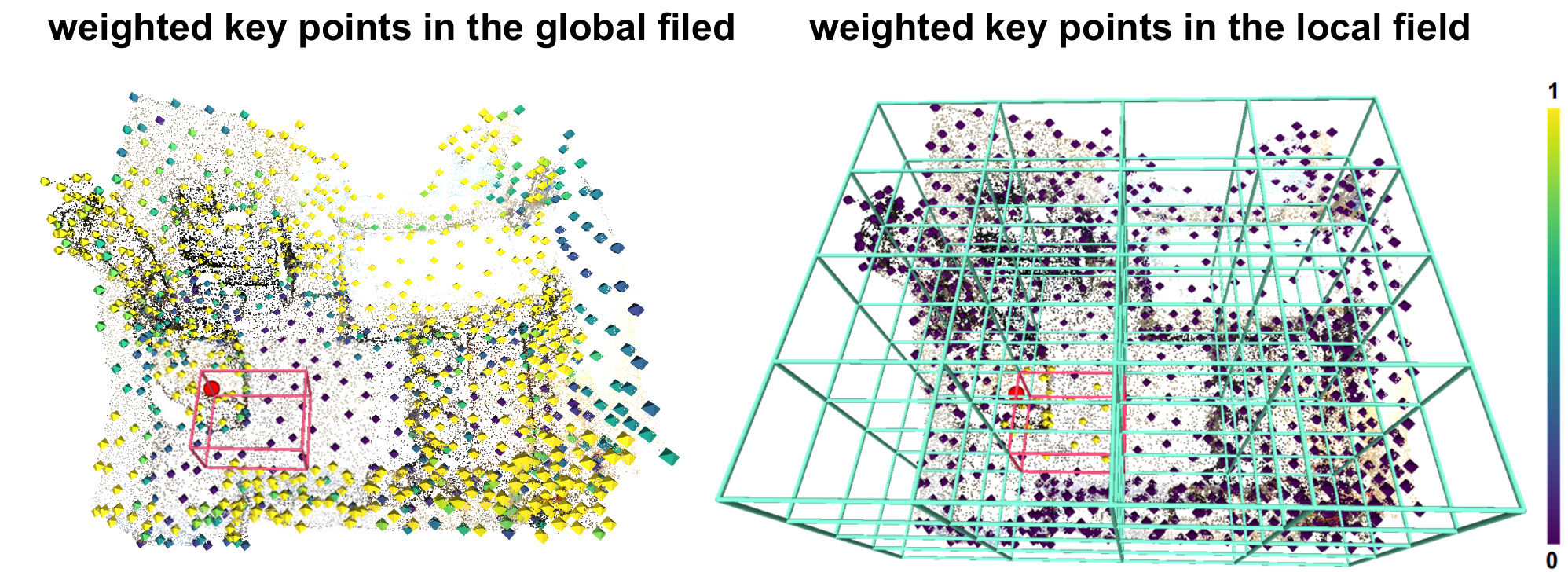}
    \caption{The visualization of the weighted key points in the two branches of SMGM. The key points are highlighted by the normalized attention weights upon the raw point clouds.}
    \label{fig:sla_vis}
\end{figure}

\subsection{Visualization}\label{sec:interesting}

\noindent\textbf{Visualization of different point sampling strategies. } We visualize the raw point clouds and the 1024 key points filtered out by different sampling strategies as shown in Fig.~\ref{fig:fps}. Sampling 1024 key points from the raw point clouds, the D-FPS shows an obvious uniform distribution in the whole space because it only considers the coordinates of the points. For the F-FPS, it shows a clear aggregation to foreground objects, since it takes the point attributes such as colors and normal vectors into account. Fusion sampling alleviates such unbalance by fusing these two strategies, but it also introduces repeated sampling points, which account for nearly 20\% of the total sampling points. Our concentration sampling shows that it further extracts a sufficient number of high-quality foreground points by redundancy removal. Also, we only take the additional attributes of the points, \ie, colors and normal vectors, as raw features without learning for F-FPS, which is highly efficient.

\begin{figure*}[t]
    \centering
    \includegraphics[height=7.8cm]{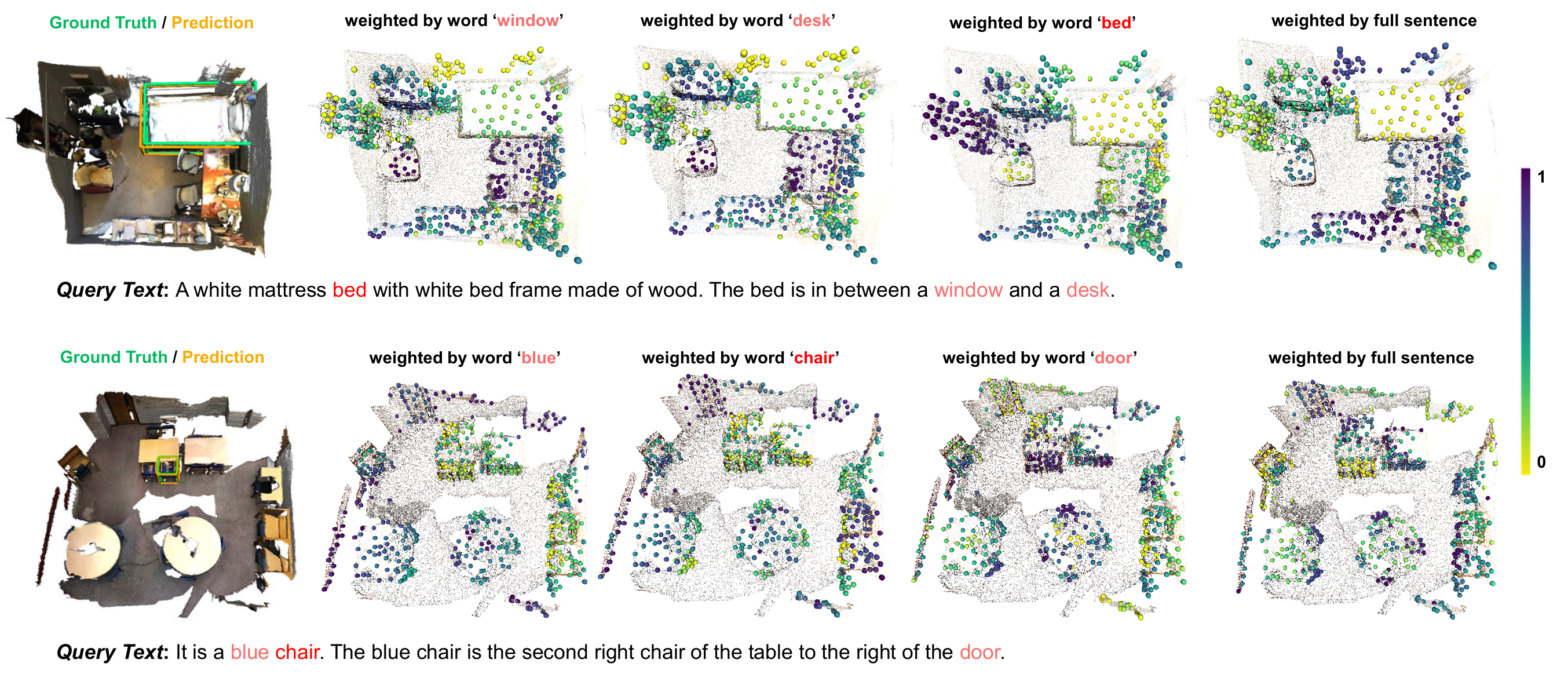}
    \caption{The visualization of the proposal points, which are attended and weighted by language representation in PLACM. The proposal points are highlighted by the normalized attention weights upon the raw point clouds.}
    \label{fig:attn}
\end{figure*}

\noindent\textbf{Visualization of the weighted key points in SMGM.} As aforementioned in Section~\ref{sec:SMGM}, two branches are designed for achieving both global and local aggregating in the cross-attention mechanism~\cite{attention-is-all-you-need}, where the proposal points act as query and the key points and language act as key and value. We visualize the key points weighted by one selected proposal point of both fields in Fig.~\ref{fig:sla_vis}, where the selected proposal point is highlighted as a red ball. We find that most of the activated key points are located on the foreground objects of the whole space in the global field. Such a mechanism encourages the proposal points to accumulate global contexts in space. However, it naturally neglects some local points inside the red box. On the contrary, for the local field, we partition the space into multiple local regions and enforce the proposal points to attend to the key points in the same region. It effectively complements local information and enables the model to accumulate global and local contexts. 

\noindent\textbf{Visualization of the proposal points weighted by multi-granularity language representation.} Remind that the proposed PLACM performs visual-linguistic interaction on both word-level and sentence-level granularities. To demonstrate the effectiveness of such a mechanism, we visualize the proposal points, which are attended by such multi-granularity language representation as shown in Fig.~\ref{fig:attn}. We can see that the normalized weights of proposal points show obvious regional distribution under different language modulations.  The first interesting observation is that the proposal points which are more relevant to the word or the sentence will be assigned a lower attention weight. We assume that such an assignment casts the proposal points into a more discriminative representation space.  We also visualize the weights by different words shown on the second column to the fourth column. One can see that not all the words will highlight their corresponding proposal points, for example, ``desk'' in the first row. We believe using partial words such as ``window'' is enough to locate the ``bed''.  For the example in the second row, the target object belongs to the challenging ``Multiple'' subset. The words ``blue'', ``chair'', ``door'' are all activated, and it finally grounds the correct ``chair'' in the scene. And the full sentence always activates the proposal points near the target object. These visualizations demonstrate that HAM can dynamically model discriminative linguistic representation for visual grounding of varying complexity.

\section{Conclusion}
\label{conclusion}
In this paper, we investigate the fundamental problems for 3D visual grounding and present an end-to-end model, namely hierarchical alignment model (HAM). We propose a novel mechanism for point-language alignment with context modulation (PLACM) to hierarchically and gradually align point clouds and language representations. We also introduce a spatially multi-granular modeling (SMGM) design to learn comprehensive spatial relationships. Additionally, we systematically analyze and incorporate three cumulative prompt engineering strategies into HAM to enhance its robustness and efficiency. 
Impressively, HAM ranked first in the ECCV 2022 ScanRefer challenge, which significantly outperformed the competing methods on both validation and testing sets. In the experiments, we present extensive ablation studies and various visualization to explain how the proposed modules work. We hope that such a simple yet effective model can inspire the research of language on 3D scenes.

\bibliographystyle{IEEEtran}
\bibliography{tpami}

\begin{thebibliography}{10}
\providecommand{\url}[1]{#1}
\csname url@samestyle\endcsname
\providecommand{\newblock}{\relax}
\providecommand{\bibinfo}[2]{#2}
\providecommand{\BIBentrySTDinterwordspacing}{\spaceskip=0pt\relax}
\providecommand{\BIBentryALTinterwordstretchfactor}{4}
\providecommand{\BIBentryALTinterwordspacing}{\spaceskip=\fontdimen2\font plus
\BIBentryALTinterwordstretchfactor\fontdimen3\font minus
  \fontdimen4\font\relax}
\providecommand{\BIBforeignlanguage}[2]{{%
\expandafter\ifx\csname l@#1\endcsname\relax
\typeout{** WARNING: IEEEtran.bst: No hyphenation pattern has been}%
\typeout{** loaded for the language `#1'. Using the pattern for}%
\typeout{** the default language instead.}%
\else
\language=\csname l@#1\endcsname
\fi
#2}}
\providecommand{\BIBdecl}{\relax}
\BIBdecl

\bibitem{scanrefer}
D.~Z. Chen, A.~X. Chang, and M.~Nie{\ss}ner, ``Scanrefer: 3d object
  localization in rgb-d scans using natural language,'' in \emph{European
  Conference on Computer Vision}.\hskip 1em plus 0.5em minus 0.4em\relax
  Springer, 2020, pp. 202--221.

\bibitem{rel3d}
A.~Goyal, K.~Yang, D.~Yang, and J.~Deng, ``Rel3d: A minimally contrastive
  benchmark for grounding spatial relations in 3d,'' \emph{Advances in Neural
  Information Processing Systems}, vol.~33, pp. 10\,514--10\,525, 2020.

\bibitem{referit3d}
P.~Achlioptas, A.~Abdelreheem, F.~Xia, M.~Elhoseiny, and L.~Guibas,
  ``Referit3d: Neural listeners for fine-grained 3d object identification in
  real-world scenes,'' in \emph{European Conference on Computer Vision}.\hskip
  1em plus 0.5em minus 0.4em\relax Springer, 2020, pp. 422--440.

\bibitem{refer-it-in-rgbd}
H.~Liu, A.~Lin, X.~Han, L.~Yang, Y.~Yu, and S.~Cui, ``Refer-it-in-rgbd: A
  bottom-up approach for 3d visual grounding in rgbd images,'' in
  \emph{Proceedings of the IEEE/CVF Conference on Computer Vision and Pattern
  Recognition}, 2021, pp. 6032--6041.

\bibitem{scannet}
A.~Dai, A.~X. Chang, M.~Savva, M.~Halber, T.~Funkhouser, and M.~Nie{\ss}ner,
  ``Scannet: Richly-annotated 3d reconstructions of indoor scenes,'' in
  \emph{Proceedings of the IEEE conference on computer vision and pattern
  recognition}, 2017, pp. 5828--5839.

\bibitem{2021mdetr}
A.~Kamath, M.~Singh, Y.~LeCun, G.~Synnaeve, I.~Misra, and N.~Carion,
  ``Mdetr-modulated detection for end-to-end multi-modal understanding,'' in
  \emph{Proceedings of the IEEE/CVF International Conference on Computer
  Vision}, 2021, pp. 1780--1790.

\bibitem{transvg}
J.~Deng, Z.~Yang, T.~Chen, W.~Zhou, and H.~Li, ``Transvg: End-to-end visual
  grounding with transformers,'' in \emph{Proceedings of the IEEE/CVF
  International Conference on Computer Vision}, 2021, pp. 1769--1779.

\bibitem{pami-grounding}
B.~A. Plummer, K.~Shih, Y.~Li, K.~Xu, S.~Lazebnik, S.~Sclaroff, and K.~Saenko,
  ``Revisiting image-language networks for open-ended phrase detection,''
  \emph{IEEE Transactions on Pattern Analysis and Machine Intelligence}, 2020.

\bibitem{TGNN}
P.-H. Huang, H.-H. Lee, H.-T. Chen, and T.-L. Liu, ``Text-guided graph neural
  networks for referring 3d instance segmentation,'' in \emph{Proceedings of
  the AAAI Conference on Artificial Intelligence}, vol.~35, 2021, pp.
  1610--1618.

\bibitem{instancerefer}
Z.~Yuan, X.~Yan, Y.~Liao, R.~Zhang, S.~Wang, Z.~Li, and S.~Cui,
  ``Instancerefer: Cooperative holistic understanding for visual grounding on
  point clouds through instance multi-level contextual referring,'' in
  \emph{Proceedings of the IEEE/CVF International Conference on Computer
  Vision}, 2021, pp. 1791--1800.

\bibitem{ffl-3dog}
M.~Feng, Z.~Li, Q.~Li, L.~Zhang, X.~Zhang, G.~Zhu, H.~Zhang, Y.~Wang, and
  A.~Mian, ``Free-form description guided 3d visual graph network for object
  grounding in point cloud,'' in \emph{Proceedings of the IEEE/CVF
  International Conference on Computer Vision}, 2021, pp. 3722--3731.

\bibitem{3dvg-transformer}
L.~Zhao, D.~Cai, L.~Sheng, and D.~Xu, ``3dvg-transformer: Relation modeling for
  visual grounding on point clouds,'' in \emph{Proceedings of the IEEE/CVF
  International Conference on Computer Vision}, 2021, pp. 2928--2937.

\bibitem{3djcg}
D.~Cai, L.~Zhao, J.~Zhang, L.~Sheng, and D.~Xu, ``3djcg: A unified framework
  for joint dense captioning and visual grounding on 3d point clouds,'' in
  \emph{Proceedings of the IEEE/CVF Conference on Computer Vision and Pattern
  Recognition}, 2022, pp. 16\,464--16\,473.

\bibitem{3dsps}
J.~Luo, J.~Fu, X.~Kong, C.~Gao, H.~Ren, H.~Shen, H.~Xia, and S.~Liu, ``3d-sps:
  Single-stage 3d visual grounding via referred point progressive selection,''
  in \emph{Proceedings of the IEEE/CVF Conference on Computer Vision and
  Pattern Recognition}, 2022, pp. 16\,454--16\,463.

\bibitem{BUTD-DETR}
A.~Jain, N.~Gkanatsios, I.~Mediratta, and K.~Fragkiadaki, ``Bottom up top down
  detection transformers for language grounding in images and point clouds,''
  in \emph{European Conference on Computer Vision (ECCV)}, 2022.

\bibitem{languagerefer}
J.~Roh, K.~Desingh, A.~Farhadi, and D.~Fox, ``Languagerefer: Spatial-language
  model for 3d visual grounding,'' in \emph{Conference on Robot
  Learning}.\hskip 1em plus 0.5em minus 0.4em\relax PMLR, 2022, pp. 1046--1056.

\bibitem{multi-view-trans}
S.~Huang, Y.~Chen, J.~Jia, and L.~Wang, ``Multi-view transformer for 3d visual
  grounding,'' in \emph{Proceedings of the IEEE/CVF Conference on Computer
  Vision and Pattern Recognition}, 2022, pp. 15\,524--15\,533.

\bibitem{sat}
Z.~Yang, S.~Zhang, L.~Wang, and J.~Luo, ``Sat: 2d semantics assisted training
  for 3d visual grounding,'' in \emph{Proceedings of the IEEE/CVF International
  Conference on Computer Vision}, 2021, pp. 1856--1866.

\bibitem{d3net}
D.~Z. Chen, Q.~Wu, M.~Nießner, and A.~X. Chang, ``D3net: A speaker-listener
  architecture for semi-supervised dense captioning and visual grounding in
  rgb-d scans,'' 2021.

\bibitem{pointnet++}
C.~R. Qi, L.~Yi, H.~Su, and L.~J. Guibas, ``Pointnet++: Deep hierarchical
  feature learning on point sets in a metric space,'' \emph{Advances in neural
  information processing systems}, vol.~30, 2017.

\bibitem{votenet}
C.~R. Qi, O.~Litany, K.~He, and L.~J. Guibas, ``Deep hough voting for 3d object
  detection in point clouds,'' in \emph{proceedings of the IEEE/CVF
  International Conference on Computer Vision}, 2019, pp. 9277--9286.

\bibitem{groupfree}
Z.~Liu, Z.~Zhang, Y.~Cao, H.~Hu, and X.~Tong, ``Group-free 3d object detection
  via transformers,'' in \emph{Proceedings of the IEEE/CVF International
  Conference on Computer Vision}, 2021, pp. 2949--2958.

\bibitem{panoptic}
A.~Kirillov, K.~He, R.~Girshick, C.~Rother, and P.~Doll{\'a}r, ``Panoptic
  segmentation,'' in \emph{Proceedings of the IEEE/CVF Conference on Computer
  Vision and Pattern Recognition}, 2019, pp. 9404--9413.

\bibitem{attention-is-all-you-need}
A.~Vaswani, N.~Shazeer, N.~Parmar, J.~Uszkoreit, L.~Jones, A.~N. Gomez,
  {\L}.~Kaiser, and I.~Polosukhin, ``Attention is all you need,''
  \emph{Advances in neural information processing systems}, vol.~30, 2017.

\bibitem{bert}
J.~Devlin, M.-W. Chang, K.~Lee, and K.~Toutanova, ``Bert: Pre-training of deep
  bidirectional transformers for language understanding,'' \emph{arXiv preprint
  arXiv:1810.04805}, 2018.

\bibitem{gpt3}
T.~Brown, B.~Mann, N.~Ryder, M.~Subbiah, J.~D. Kaplan, P.~Dhariwal,
  A.~Neelakantan, P.~Shyam, G.~Sastry, A.~Askell \emph{et~al.}, ``Language
  models are few-shot learners,'' \emph{Advances in neural information
  processing systems}, vol.~33, pp. 1877--1901, 2020.

\bibitem{bridge-prompt}
M.~Li, L.~Chen, Y.~Duan, Z.~Hu, J.~Feng, J.~Zhou, and J.~Lu, ``Bridge-prompt:
  Towards ordinal action understanding in instructional videos,'' in
  \emph{Proceedings of the IEEE/CVF Conference on Computer Vision and Pattern
  Recognition}, 2022, pp. 19\,880--19\,889.

\bibitem{vqa}
S.~Antol, A.~Agrawal, J.~Lu, M.~Mitchell, D.~Batra, C.~L. Zitnick, and
  D.~Parikh, ``Vqa: Visual question answering,'' in \emph{Proceedings of the
  IEEE international conference on computer vision}, 2015, pp. 2425--2433.

\bibitem{vqav2}
Y.~Goyal, T.~Khot, D.~Summers-Stay, D.~Batra, and D.~Parikh, ``Making the v in
  vqa matter: Elevating the role of image understanding in visual question
  answering,'' in \emph{Proceedings of the IEEE conference on computer vision
  and pattern recognition}, 2017, pp. 6904--6913.

\bibitem{visual-dialog}
A.~Das, S.~Kottur, K.~Gupta, A.~Singh, D.~Yadav, J.~M. Moura, D.~Parikh, and
  D.~Batra, ``Visual dialog,'' in \emph{Proceedings of the IEEE conference on
  computer vision and pattern recognition}, 2017, pp. 326--335.

\bibitem{wstg_compose}
J.~Chen, W.~Luo, W.~Zhang, and L.~Ma, ``Explore inter-contrast between videos
  via composition for weakly supervised temporal sentence grounding,''
  \emph{Proceedings of the AAAI Conference on Artificial Intelligence},
  vol.~36, no.~1, pp. 267--275, Jun. 2022.

\bibitem{video-cap}
W.~Zhang, B.~Wang, L.~Ma, and W.~Liu, ``Reconstruct and represent video
  contents for captioning via reinforcement learning,'' \emph{IEEE transactions
  on pattern analysis and machine intelligence}, vol.~42, no.~12, pp.
  3088--3101, 2019.

\bibitem{flickr}
P.~Young, A.~Lai, M.~Hodosh, and J.~Hockenmaier, ``From image descriptions to
  visual denotations: New similarity metrics for semantic inference over event
  descriptions,'' \emph{Transactions of the Association for Computational
  Linguistics}, vol.~2, pp. 67--78, 2014.

\bibitem{scan2cap}
Z.~Chen, A.~Gholami, M.~Nie{\ss}ner, and A.~X. Chang, ``Scan2cap: Context-aware
  dense captioning in rgb-d scans,'' in \emph{Proceedings of the IEEE/CVF
  Conference on Computer Vision and Pattern Recognition}, 2021, pp. 3193--3203.

\bibitem{dalle}
A.~Ramesh, M.~Pavlov, G.~Goh, S.~Gray, C.~Voss, A.~Radford, M.~Chen, and
  I.~Sutskever, ``Zero-shot text-to-image generation,'' in \emph{International
  Conference on Machine Learning}.\hskip 1em plus 0.5em minus 0.4em\relax PMLR,
  2021, pp. 8821--8831.

\bibitem{diffusion}
R.~Rombach, A.~Blattmann, D.~Lorenz, P.~Esser, and B.~Ommer, ``High-resolution
  image synthesis with latent diffusion models,'' 2021.

\bibitem{vl-navigation}
Y.~Qi, Z.~Pan, S.~Zhang, A.~v.~d. Hengel, and Q.~Wu, ``Object-and-action aware
  model for visual language navigation,'' in \emph{European Conference on
  Computer Vision}.\hskip 1em plus 0.5em minus 0.4em\relax Springer, 2020, pp.
  303--317.

\bibitem{embodied-qa}
A.~Das, S.~Datta, G.~Gkioxari, S.~Lee, D.~Parikh, and D.~Batra, ``Embodied
  question answering,'' in \emph{Proceedings of the IEEE Conference on Computer
  Vision and Pattern Recognition}, 2018, pp. 1--10.

\bibitem{referitgame}
S.~Kazemzadeh, V.~Ordonez, M.~Matten, and T.~Berg, ``Referitgame: Referring to
  objects in photographs of natural scenes,'' in \emph{Proceedings of the 2014
  conference on empirical methods in natural language processing (EMNLP)},
  2014, pp. 787--798.

\bibitem{plummer2015flickr30k}
B.~A. Plummer, L.~Wang, C.~M. Cervantes, J.~C. Caicedo, J.~Hockenmaier, and
  S.~Lazebnik, ``Flickr30k entities: Collecting region-to-phrase
  correspondences for richer image-to-sentence models,'' in \emph{Proceedings
  of the IEEE international conference on computer vision}, 2015, pp.
  2641--2649.

\bibitem{mao2016generation}
J.~Mao, J.~Huang, A.~Toshev, O.~Camburu, A.~L. Yuille, and K.~Murphy,
  ``Generation and comprehension of unambiguous object descriptions,'' in
  \emph{Proceedings of the IEEE conference on computer vision and pattern
  recognition}, 2016, pp. 11--20.

\bibitem{yu2016modeling}
L.~Yu, P.~Poirson, S.~Yang, A.~C. Berg, and T.~L. Berg, ``Modeling context in
  referring expressions,'' in \emph{European Conference on Computer
  Vision}.\hskip 1em plus 0.5em minus 0.4em\relax Springer, 2016, pp. 69--85.

\bibitem{yu2018mattnet}
L.~Yu, Z.~Lin, X.~Shen, J.~Yang, X.~Lu, M.~Bansal, and T.~L. Berg, ``Mattnet:
  Modular attention network for referring expression comprehension,'' in
  \emph{Proceedings of the IEEE Conference on Computer Vision and Pattern
  Recognition}, 2018, pp. 1307--1315.

\bibitem{wang2019neighbourhood}
P.~Wang, Q.~Wu, J.~Cao, C.~Shen, L.~Gao, and A.~v.~d. Hengel, ``Neighbourhood
  watch: Referring expression comprehension via language-guided graph attention
  networks,'' in \emph{Proceedings of the IEEE/CVF Conference on Computer
  Vision and Pattern Recognition}, 2019, pp. 1960--1968.

\bibitem{yang-osvg1}
Z.~Yang, B.~Gong, L.~Wang, W.~Huang, D.~Yu, and J.~Luo, ``A fast and accurate
  one-stage approach to visual grounding,'' in \emph{Proceedings of the
  IEEE/CVF International Conference on Computer Vision}, 2019, pp. 4683--4693.

\bibitem{yang-osvg2}
Z.~Yang, T.~Chen, L.~Wang, and J.~Luo, ``Improving one-stage visual grounding
  by recursive sub-query construction,'' in \emph{European Conference on
  Computer Vision}.\hskip 1em plus 0.5em minus 0.4em\relax Springer, 2020, pp.
  387--404.

\bibitem{huang2021look}
B.~Huang, D.~Lian, W.~Luo, and S.~Gao, ``Look before you leap: Learning
  landmark features for one-stage visual grounding,'' in \emph{Proceedings of
  the IEEE/CVF Conference on Computer Vision and Pattern Recognition}, 2021,
  pp. 16\,888--16\,897.

\bibitem{clip}
A.~Radford, J.~W. Kim, C.~Hallacy, A.~Ramesh, G.~Goh, S.~Agarwal, G.~Sastry,
  A.~Askell, P.~Mishkin, J.~Clark \emph{et~al.}, ``Learning transferable visual
  models from natural language supervision,'' in \emph{International Conference
  on Machine Learning}.\hskip 1em plus 0.5em minus 0.4em\relax PMLR, 2021, pp.
  8748--8763.

\bibitem{transrefer3d}
D.~He, Y.~Zhao, J.~Luo, T.~Hui, S.~Huang, A.~Zhang, and S.~Liu, ``Transrefer3d:
  Entity-and-relation aware transformer for fine-grained 3d visual grounding,''
  in \emph{Proceedings of the 29th ACM International Conference on Multimedia},
  2021, pp. 2344--2352.

\bibitem{birdview1}
J.~Ku, M.~Mozifian, J.~Lee, A.~Harakeh, and S.~L. Waslander, ``Joint 3d
  proposal generation and object detection from view aggregation,'' in
  \emph{2018 IEEE/RSJ International Conference on Intelligent Robots and
  Systems (IROS)}.\hskip 1em plus 0.5em minus 0.4em\relax IEEE, 2018, pp. 1--8.

\bibitem{birdview2}
M.~Liang, B.~Yang, S.~Wang, and R.~Urtasun, ``Deep continuous fusion for
  multi-sensor 3d object detection,'' in \emph{Proceedings of the European
  conference on computer vision (ECCV)}, 2018, pp. 641--656.

\bibitem{birdview3}
B.~Yang, W.~Luo, and R.~Urtasun, ``Pixor: Real-time 3d object detection from
  point clouds,'' in \emph{Proceedings of the IEEE conference on Computer
  Vision and Pattern Recognition}, 2018, pp. 7652--7660.

\bibitem{frontalview1}
X.~Chen, H.~Ma, J.~Wan, B.~Li, and T.~Xia, ``Multi-view 3d object detection
  network for autonomous driving,'' in \emph{Proceedings of the IEEE conference
  on Computer Vision and Pattern Recognition}, 2017, pp. 1907--1915.

\bibitem{frontalview2}
B.~Xu and Z.~Chen, ``Multi-level fusion based 3d object detection from
  monocular images,'' in \emph{Proceedings of the IEEE conference on computer
  vision and pattern recognition}, 2018, pp. 2345--2353.

\bibitem{voxel1}
S.~Song and J.~Xiao, ``Deep sliding shapes for amodal 3d object detection in
  rgb-d images,'' in \emph{Proceedings of the IEEE conference on computer
  vision and pattern recognition}, 2016, pp. 808--816.

\bibitem{voxel2}
Y.~Zhou and O.~Tuzel, ``Voxelnet: End-to-end learning for point cloud based 3d
  object detection,'' in \emph{Proceedings of the IEEE conference on computer
  vision and pattern recognition}, 2018, pp. 4490--4499.

\bibitem{frustum-pointnets}
C.~R. Qi, W.~Liu, C.~Wu, H.~Su, and L.~J. Guibas, ``Frustum pointnets for 3d
  object detection from rgb-d data,'' in \emph{Proceedings of the IEEE
  conference on computer vision and pattern recognition}, 2018, pp. 918--927.

\bibitem{pointrcnn}
S.~Shi, X.~Wang, and H.~Li, ``Pointrcnn: 3d object proposal generation and
  detection from point cloud,'' in \emph{Proceedings of the IEEE/CVF conference
  on computer vision and pattern recognition}, 2019, pp. 770--779.

\bibitem{RepSurf-U}
H.~Ran, J.~Liu, and C.~Wang, ``Surface representation for point clouds,'' in
  \emph{Proceedings of the IEEE/CVF Conference on Computer Vision and Pattern
  Recognition}, 2022, pp. 18\,942--18\,952.

\bibitem{FCAF3D}
D.~Rukhovich, A.~Vorontsova, and A.~Konushin, ``Fcaf3d: Fully convolutional
  anchor-free 3d object detection,'' \emph{arXiv preprint arXiv:2112.00322},
  2021.

\bibitem{sun-rgbd}
S.~Song, S.~P. Lichtenberg, and J.~Xiao, ``Sun rgb-d: A rgb-d scene
  understanding benchmark suite,'' in \emph{Proceedings of the IEEE conference
  on computer vision and pattern recognition}, 2015, pp. 567--576.

\bibitem{S3DIS}
I.~Armeni, O.~Sener, A.~R. Zamir, H.~Jiang, I.~Brilakis, M.~Fischer, and
  S.~Savarese, ``3d semantic parsing of large-scale indoor spaces,'' in
  \emph{Proceedings of the IEEE conference on computer vision and pattern
  recognition}, 2016, pp. 1534--1543.

\bibitem{Lian_2022_SSF}
D.~Lian, D.~Zhou, J.~Feng, and X.~Wang, ``Scaling \& shifting your features: A
  new baseline for efficient model tuning,'' in \emph{Advances in Neural
  Information Processing Systems (NeurIPS)}, 2022.

\bibitem{cao2022circular}
S.~Cao, W.~Luo, B.~Wang, W.~Zhang, and L.~Ma, ``A circular window-based cascade
  transformer for online action detection,'' \emph{arXiv preprint
  arXiv:2208.14209}, 2022.

\bibitem{qian2022svip}
Y.~Qian, W.~Luo, D.~Lian, X.~Tang, P.~Zhao, and S.~Gao, ``Svip: Sequence
  verification for procedures in videos,'' in \emph{Proceedings of the IEEE/CVF
  Conference on Computer Vision and Pattern Recognition}, 2022, pp.
  19\,890--19\,902.

\bibitem{devlin2018bert}
J.~Devlin, M.-W. Chang, K.~Lee, and K.~Toutanova, ``Bert: Pre-training of deep
  bidirectional transformers for language understanding,'' \emph{arXiv preprint
  arXiv:1810.04805}, 2018.

\bibitem{roberta}
Y.~Liu, M.~Ott, N.~Goyal, J.~Du, M.~Joshi, D.~Chen, O.~Levy, M.~Lewis,
  L.~Zettlemoyer, and V.~Stoyanov, ``Roberta: A robustly optimized bert
  pretraining approach,'' \emph{arXiv preprint arXiv:1907.11692}, 2019.

\bibitem{tcl}
J.~Yang, J.~Duan, S.~Tran, Y.~Xu, S.~Chanda, L.~Chen, B.~Zeng, T.~Chilimbi, and
  J.~Huang, ``Vision-language pre-training with triple contrastive learning,''
  in \emph{Proceedings of the IEEE/CVF Conference on Computer Vision and
  Pattern Recognition}, 2022, pp. 15\,671--15\,680.

\bibitem{3dssd}
Z.~Yang, Y.~Sun, S.~Liu, and J.~Jia, ``3dssd: Point-based 3d single stage
  object detector,'' in \emph{Proceedings of the IEEE/CVF conference on
  computer vision and pattern recognition}, 2020, pp. 11\,040--11\,048.

\bibitem{glove}
J.~Pennington, R.~Socher, and C.~D. Manning, ``Glove: Global vectors for word
  representation,'' in \emph{Proceedings of the 2014 conference on empirical
  methods in natural language processing (EMNLP)}, 2014, pp. 1532--1543.

\bibitem{gru}
J.~Chung, C.~Gulcehre, K.~Cho, and Y.~Bengio, ``Empirical evaluation of gated
  recurrent neural networks on sequence modeling,'' \emph{arXiv preprint
  arXiv:1412.3555}, 2014.

\bibitem{distilbert}
V.~Sanh, L.~Debut, J.~Chaumond, and T.~Wolf, ``Distilbert, a distilled version
  of bert: smaller, faster, cheaper and lighter,'' \emph{arXiv preprint
  arXiv:1910.01108}, 2019.

\bibitem{explore}
J.~Chen, W.~Luo, W.~Zhang, and L.~Ma, ``Explore inter-contrast between videos
  via composition for weakly supervised temporal sentence grounding,'' in
  \emph{Proceedings of the AAAI Conference on Artificial Intelligence},
  vol.~36, no.~1, 2022, pp. 267--275.

\bibitem{vilt}
W.~Kim, B.~Son, and I.~Kim, ``Vilt: Vision-and-language transformer without
  convolution or region supervision,'' in \emph{International Conference on
  Machine Learning}.\hskip 1em plus 0.5em minus 0.4em\relax PMLR, 2021, pp.
  5583--5594.

\bibitem{swin-transformer}
Z.~Liu, Y.~Lin, Y.~Cao, H.~Hu, Y.~Wei, Z.~Zhang, S.~Lin, and B.~Guo, ``Swin
  transformer: Hierarchical vision transformer using shifted windows,'' in
  \emph{Proceedings of the IEEE/CVF International Conference on Computer
  Vision}, 2021, pp. 10\,012--10\,022.

\bibitem{ViT}
A.~Dosovitskiy, L.~Beyer, A.~Kolesnikov, D.~Weissenborn, X.~Zhai,
  T.~Unterthiner, M.~Dehghani, M.~Minderer, G.~Heigold, S.~Gelly, J.~Uszkoreit,
  and N.~Houlsby, ``An image is worth 16x16 words: Transformers for image
  recognition at scale,'' 2021.

\bibitem{beauty-detr}
A.~Jain, N.~Gkanatsios, I.~Mediratta, and K.~Fragkiadaki, ``Looking outside the
  box to ground language in 3d scenes,'' \emph{arXiv preprint
  arXiv:2112.08879}, 2021.

\bibitem{adamw}
I.~Loshchilov and F.~Hutter, ``Decoupled weight decay regularization,''
  \emph{arXiv preprint arXiv:1711.05101}, 2017.

\bibitem{lstm}
S.~Hochreiter and J.~Schmidhuber, ``Long short-term memory,'' \emph{Neural
  computation}, vol.~9, no.~8, pp. 1735--1780, 1997.

\bibitem{enet}
A.~Paszke, A.~Chaurasia, S.~Kim, and E.~Culurciello, ``Enet: A deep neural
  network architecture for real-time semantic segmentation,'' \emph{arXiv
  preprint arXiv:1606.02147}, 2016.

\end{thebibliography}

\end{document}